%% file: main.tex
\documentclass[10pt,twocolumn,letterpaper]{article}

\usepackage{cvpr}              %

\input{preamble}
\definecolor{cvprblue}{rgb}{0.21,0.49,0.74}
\usepackage[pagebackref,breaklinks,colorlinks,allcolors=cvprblue]{hyperref}
\usepackage{booktabs}
\usepackage{mathrsfs}
\usepackage{float} 
\usepackage{colortbl}
\PassOptionsToPackage{table}{xcolor}

\def\paperID{*****} %
\def\confName{CVPR}
\def\confYear{2025}

\definecolor{somegray}{rgb}{0.5, 0.5, 0.5}
\newcommand{\darkgrayed}[1]{\textcolor{somegray}{#1}}
\makeatletter
\newcommand*\titleheader[1]{\gdef\@titleheader{#1}}
\AtBeginDocument{%
  \let\st@red@title\@title
  \def\@title{%
    \vskip-3em
    \bgroup\normalfont\small\centering\@titleheader\par\egroup
    \vskip1.5em\st@red@title}
}
\makeatother

\titleheader{\darkgrayed{This paper has been accepted for publication at the\\
IEEE Computer Vision and Pattern Recognition Workshop (CVPRW), Nashville, 2025.
\copyright IEEE}}

\title{Egocentric Event-Based Vision for Ping Pong Ball Trajectory Prediction}

\author{Ivan Alberico \hspace{3mm} Marco Cannici \hspace{3mm} Giovanni Cioffi \hspace{3mm} Davide Scaramuzza\\
Robotics and Perception Group, University of Zurich, Switzerland\\
{\tt\small }
}

\begin{document}

\maketitle
\input{sec/0_abstract}

\input{sec/8_supplementary_material}
\input{sec/1_intro}
\input{sec/2_related_works}

\input{sec/3_method}

\input{sec/4_latency_analysis}
\input{sec/5_experiments}

\input{sec/6_discussion}

\input{sec/7_conclusions}
{
    \small
    \bibliographystyle{ieeenat_fullname}
    \bibliography{main}
}

\input{sec/X_suppl}

\end{document}

%% file: sec/0_abstract.tex
\begin{abstract}

In this paper, we present a real-time egocentric trajectory prediction system for table tennis using event cameras. Unlike standard cameras, which suffer from high latency and motion blur at fast ball speeds, event cameras provide higher temporal resolution, allowing more frequent state updates, greater robustness to outliers, and accurate trajectory predictions using just a short time window after the opponent's impact. We collect a dataset of ping-pong game sequences, including 3D ground-truth trajectories of the ball, synchronized with sensor data from the Meta Project Aria glasses and event streams.  Our system leverages foveated vision, using eye-gaze data from the glasses to process only events in the viewer’s fovea. This biologically inspired approach improves ball detection performance and significantly reduces computational latency, as it efficiently allocates resources to the most perceptually relevant regions, achieving a reduction factor of \textbf{10.81} on the collected trajectories. 
Our detection pipeline has a worst-case total latency of 4.5 ms, including computation and perception--significantly lower than a frame-based 30 FPS system, which, in the worst case, takes 66 ms solely for perception.
Finally, we fit a trajectory prediction model to the estimated states of the ball, enabling 3D trajectory forecasting in the future. To the best of our knowledge, this is the first approach to predict table tennis trajectories from an egocentric perspective using event cameras.

\end{abstract}

%% file: sec/8_supplementary_material.tex
\section*{Supplementary Material}\label{sec:SupplementaryMaterial}

{\small \textbf{Code}: \url{https://github.com/uzh-rpg/event_based_ping_pong_ball_trajectory_prediction}}

%% file: sec/1_intro.tex
\section{Introduction} \label{sec:intro}

In recent years, the task of tracking fast-moving objects like a ping pong ball in real-time has gained increasing attention, particularly for applications in AR/VR gaming~\cite{Ma_2024}~\cite{wang2024strategyskilllearningphysicsbased}, real-time sports analysis~\cite{Zhu2024}, and robotics~\cite{D_Ambrosio_2023}~\cite{9892776}~\cite{Tebbe2018ATT}~\cite{dambrosio2024achievinghumanlevelcompetitive}. The challenge lies in the precise perception required to track these objects as they move at high speeds (in top players even reaching $20$ to $30$ m/s \cite{schneider2022table}), while minimizing the associated bandwidth costs and sensing latency. 

Traditional tracking systems typically rely on frame-based, high-resolution cameras that operate at extremely high frame rates (e.g., 120–600 FPS in table tennis applications~\cite{TTNet, dambrosio2024achievinghumanlevelcompetitive_long, achterhold2023blackbox, 6288166, Gossard_2023, Tamaki2004MeasuringBS}). Although this approach is effective, it comes with the drawback of consuming substantial bandwidth and computational power. This results in a fundamental trade-off between latency, bandwidth, and accuracy, making it crucial to balance the need for real-time performance with the constraints of system resources. While existing frame-based vision systems have successfully achieved real-time performance using fixed cameras~\cite{TTNet} \cite{8957482} \cite{10209039}, none have been adapted to an egocentric perspective: that of the player.

\begin{figure}[t]
    \centering
    \begin{subfigure}{0.474\linewidth}
        \centering
        \includegraphics[width=\linewidth]{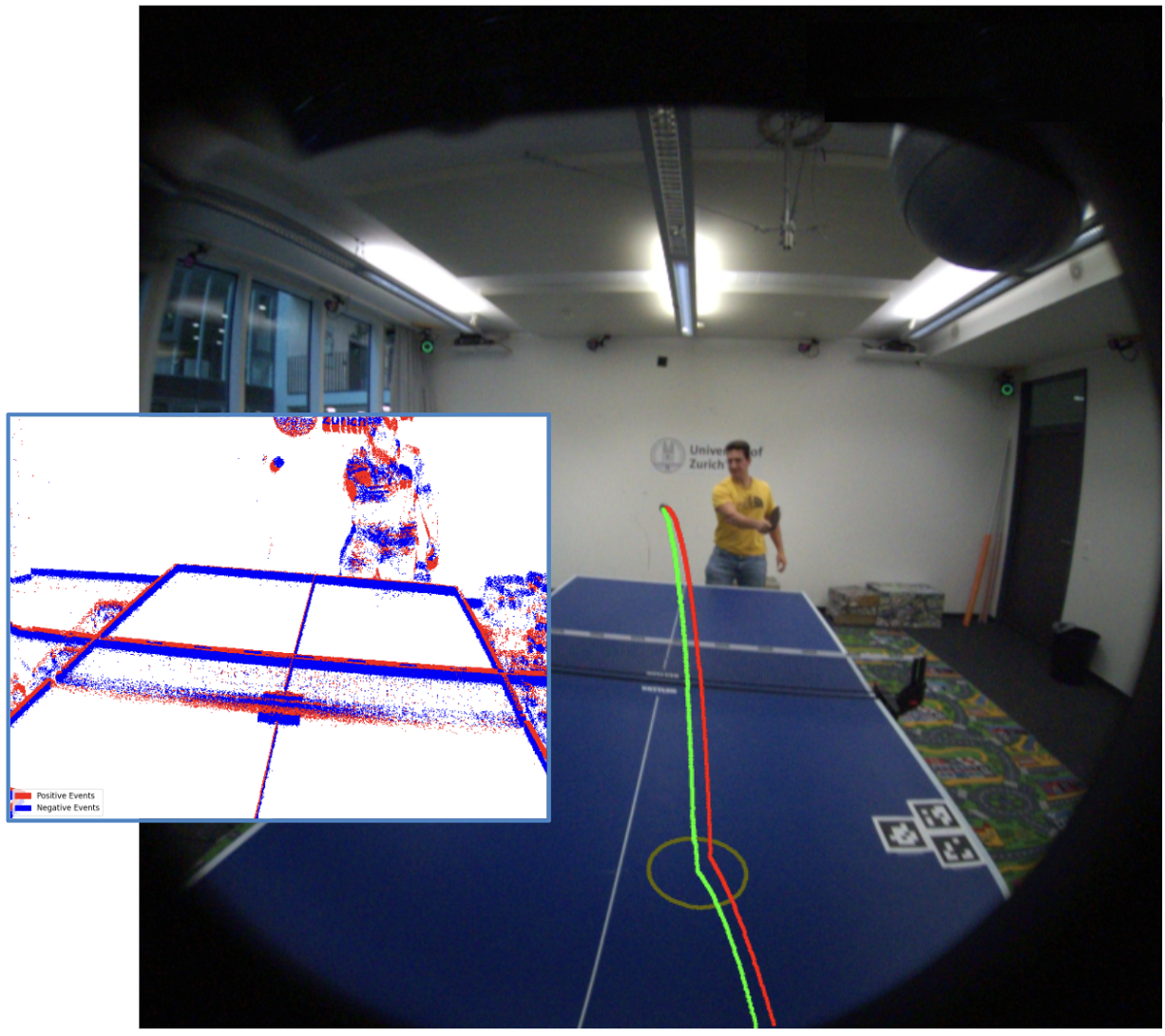}
    \end{subfigure}
    \hfill
    \begin{subfigure}{0.49\linewidth}
        \centering
        \includegraphics[width=\linewidth]{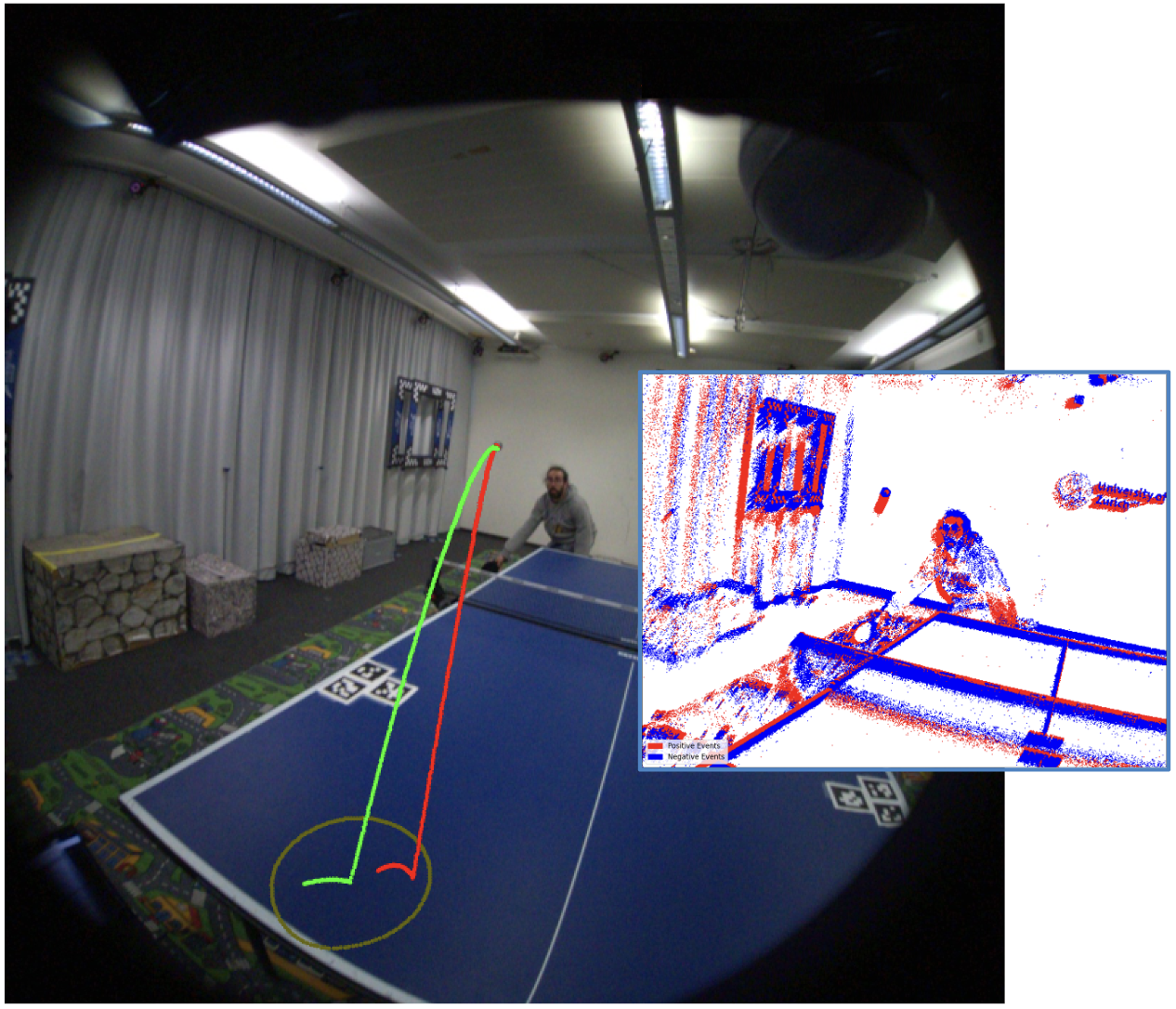}
    \end{subfigure}
    \caption{Visual representation of trajectory prediction reprojected on the Aria RGB camera frame. The green line indicates the ground truth trajectory, while the red line the predicted trajectory. 
    }
    \label{fig:prediction_sequences}
\end{figure}

Meta recently introduced Project Aria~\cite{engel2023projectarianewtool}, an experimental research initiative consisting of a smart-glasses device designed to explore how AI-driven AR can enhance real-world interactions. Equipped with an array of sensors—including cameras~\cite{goesele25imaging}, inertial measurement units (IMUs), and microphones—Project Aria glasses capture video and audio, as well as eye tracking and location data. A key feature of the system is its eye-gaze tracking module, which provides real-time gaze data that can be leveraged to implement foveated vision. A system can prioritize and process only the most relevant visual information from the foveal region around the gaze, helping to balance the trade-off between latency and accuracy while reducing sensor bandwidth. On the other hand, a limitation of these glasses is that they record videos at a maximum frame rate of 30 FPS, which may not be sufficient for high-speed scenarios like table tennis or other sports where objects move at extreme velocities. In these cases, the lower frame rate could hinder real-time capabilities, making it challenging to capture rapid motion with the required precision.

Event cameras, a neuromorphic sensing technology, offer a promising alternative to conventional frame-based cameras for high-speed, dynamic applications. Unlike traditional cameras, which capture frames at fixed intervals, event cameras operate asynchronously, detecting changes in brightness at individual pixels. This allows them to achieve extremely high temporal resolution in the order of microseconds while avoiding motion blur and reducing bandwidth~\cite{Gallego_2022}. 
While previous research has demonstrated the potential of event cameras for tasks like estimating the spin of ping pong balls~\cite{Gossard2024TableTB}~\cite{Nakabayashi_2024_CVPR}, these studies rely again on fixed camera systems, such as lateral or top-down views, where the ball moves most of the times without occlusions and on a static background. However, deploying tracking systems in dynamic, egocentric scenarios, such as those enabled by devices like Project Aria, introduces new challenges like the difficulty of isolating the ball due to the background movement of the opponents or obtaining an accurate estimate of the trajectory under small parallax angles.

\begin{figure}[t]
     \vspace{2mm}
     \centering
     \includegraphics[width=0.30\textwidth]{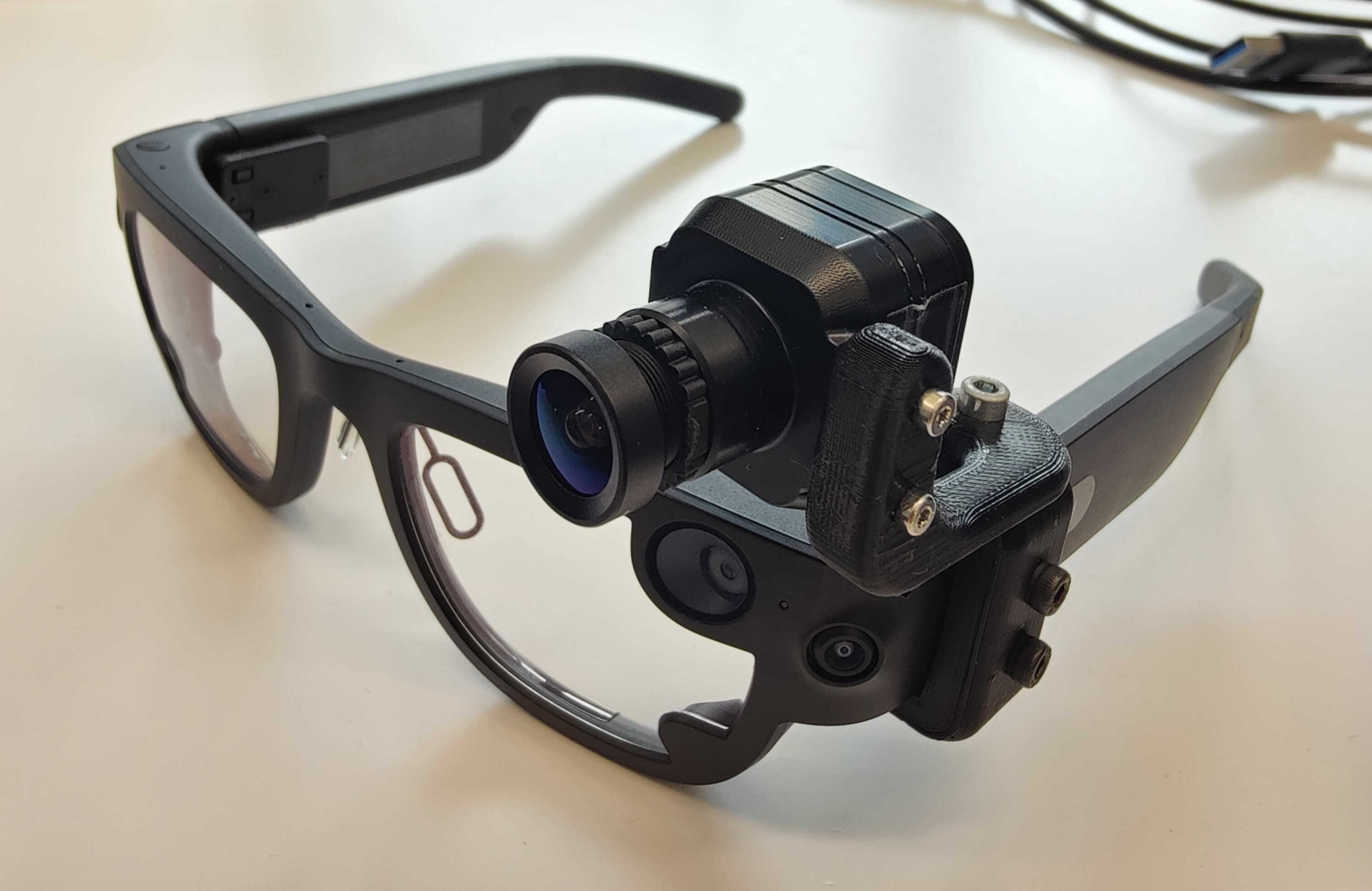}
     \caption{Configuration of the Meta Project Aria glasses with an event camera mounted on top, aligned with the built-in RGB camera to maximize the overlap between their fields of view.}
     \label{img:aria_dvs_setup}
\end{figure}

In this work, we address the problem of egocentric view trajectory prediction of a table-tennis ball by leveraging the unique capabilities of event cameras, with the setup shown in Figure~\ref{img:aria_dvs_setup}. Our method uses foveated vision to crop within a region around the eye-gaze reprojection; it then applies motion compensation~\cite{falanga2020dynamic} to distinguish moving objects from static ones. 
Finally, it predicts the ball trajectory by fitting multiple measurements collected over a time window. Our contributions are summarized as follows:
\begin{itemize}
    \item We present the first framework for egocentric table-tennis ball trajectory prediction using event cameras.
    \item We present a dataset with 3D ground-truth ball trajectories synchronized with multi-modal sensor data from Meta Project Aria glasses and event cameras.
    \item We demonstrate that event-based algorithms can capture significantly more measurements within the same time window compared to frame-based cameras, with our system operating at 200 Hz, whereas a traditional setup using Project Aria glasses runs at only 30 Hz, leading to improved performance.
    When using traditional physics-based trajectory prediction, the higher measurement frequency of the event-based pipeline reduces the average error by $4.8$ cm compared to frame-based updates over a $0.2$ s time horizon. With learning-based prediction methods~\cite{8957482}, this error reduction further improves to $8.4$ cm.
    \item We conduct a latency analysis of our algorithm, highlighting the latency benefits of event camera-based algorithms when combined with the eye-tracker output of the Project Aria glasses. Our method achieves a computation latency of just $1.5$ ms for reliably detecting the ball. This leads to an ideal worst-case total latency \cite{forrai2023event} of just $4.5$ ms, which is lower than that of a $30$ FPS camera, where perception alone, excluding computation, takes $66$ ms.
    
\end{itemize}

%% file: sec/2_related_works.tex
\section{Related Works} \label{sec:related_works}
Fast-paced sports like table tennis require precise, low-latency sensing for accurate tracking and prediction.
Traditional methods often struggle with motion blur and high latency, especially under dynamic conditions, which has motivated the exploration of event cameras as an alternative sensing modality. These cameras have demonstrated significant potential in the context of tracking dynamic obstacles. 
Event-based approaches like SpikeMOT~\cite{Wang_2025}, EVtracker~\cite{s22166090}, and~\cite{Perez_Salesa_2022} have demonstrated robust motion blur-resistant tracking in dynamic environments.
Additionally, some works have specifically addressed ball detection in cluttered scenes~\cite{7759345}, which leverage spatiotemporal event data to improve detection accuracy and maintain focus on relevant objects despite background distractions.

Beyond tracking, event cameras have also proven valuable for high-speed obstacle avoidance. \cite{falanga2019howfast} analyzed the impact of perception latency on a robot's maximum safe navigation speed, highlighting that lower latency sensors like event cameras can enhance high-speed obstacle avoidance capabilities. Building on this, a framework was developed to allow quadrotors to dodge fast-moving objects using onboard event cameras~\cite{falanga2020dynamic}. Similarly, while shallow neural networks have been explored for obstacle avoidance~\cite{wang2022Fast}, they often face higher sensing latency, limiting their ability to respond quickly. Event cameras have also been successfully used in high-speed robotic ball catching. For instance, ~\cite{forrai2023event} presents the first successful demonstration of a quadrupedal robot equipped with a net catching an object with an event camera. In a similar fashion, EV-Catcher~\cite{wang2022Fast} presents a static setup for ping-pong ball catching, exemplifying how event-based neural networks process asynchronous data in real time, allowing for precise and rapid responses to fast-moving objects. These advancements highlight the potential of event cameras in applications demanding swift reactions to dynamic obstacles. 

Trajectory prediction has been a fundamental research area in sports robotics, particularly for table tennis. Various studies have addressed the task of estimating the ball's trajectory during gameplay. For instance, TTNet~\cite{TTNet} introduces a neural network model for real-time table tennis video analysis, enabling event detection, ball tracking, and segmentation. Other neural network-based approaches, such as graph neural networks~\cite{GNN_Tianjian24} or deep conditional generative models~\cite{8957482}, have also been employed to enhance the accuracy and robustness of trajectory tracking and prediction for robotic table tennis systems. Alternatively,~\cite{achterhold2023blackbox} follows a grey-box approach, combining a physical model with data-driven learning to filter and predict table tennis ball trajectories using an Extended Kalman Filter and a neural model for estimating initial conditions. While all these methods rely on frame-based solutions to estimate the trajectory of the ping pong ball, we aim to propose a fully event-based pipeline that takes advantage of the low-latency, low-bandwidth capabilities of event cameras.

In addition to trajectory prediction, researchers have focused extensively on estimating ball spin, as it significantly affects the flight path and rebound behavior of a table tennis ball. Spin estimation has been addressed through various approaches, ranging from image registration techniques~\cite{Tamaki2004MeasuringBS}~\cite{6288166} to specialized spin detection algorithms using deep learning networks~\cite{Gossard_2023} or more traditional methods~\cite{10285571}~\cite{9196536}. For example, some methods have employed asynchronous cameras~\cite{Tamaki20243753} to measure spin without the need for synchronized shutters or high-speed cameras, while others have used quaternion-based filters to track spin dynamics~\cite{6907460}. Building on the importance of spin estimation, \cite{9467347} develops a deep reinforcement learning approach to learn a ball stroke strategy by incorporating spin velocity estimation. Spin detection has proven invaluable for robotic systems aiming to return strokes effectively, as accurate spin information allows for precise trajectory adjustments.

Event cameras have been investigated for spin estimation~\cite{Gossard2024TableTB}~\cite{Nakabayashi_2024_CVPR}.
Recently, \cite{Ziegler2025arxiv} introduced a real-time table tennis robot perception pipeline using a stereo event camera setup, achieving higher update rates and improved ball position, velocity, and spin estimation with reduced errors compared to frame-based approaches. Analogously, \cite{10209039} presented a fast trajectory end-point prediction method using event cameras and an LSTM-based model to leverage temporal event data in reactive robot control. However, all these prior works were conducted exclusively in static setups, where the ball’s motion was constrained or externally controlled. While these efforts demonstrate the potential of event cameras for high-speed spin estimation, their applicability to egocentric gameplay scenarios remains unexplored. In our work, we do not focus on estimating the ball's spin, as doing so from the player's perspective would be challenging, even with an event camera. Instead, we concentrate on providing low-latency tracking of the ball's state, enabling us to predict its trajectory from an egocentric view.

%% file: sec/3_method.tex
\section{Method} \label{sec:method}
\begin{figure}[t]
     \centering
     \includegraphics[width=0.45\textwidth]{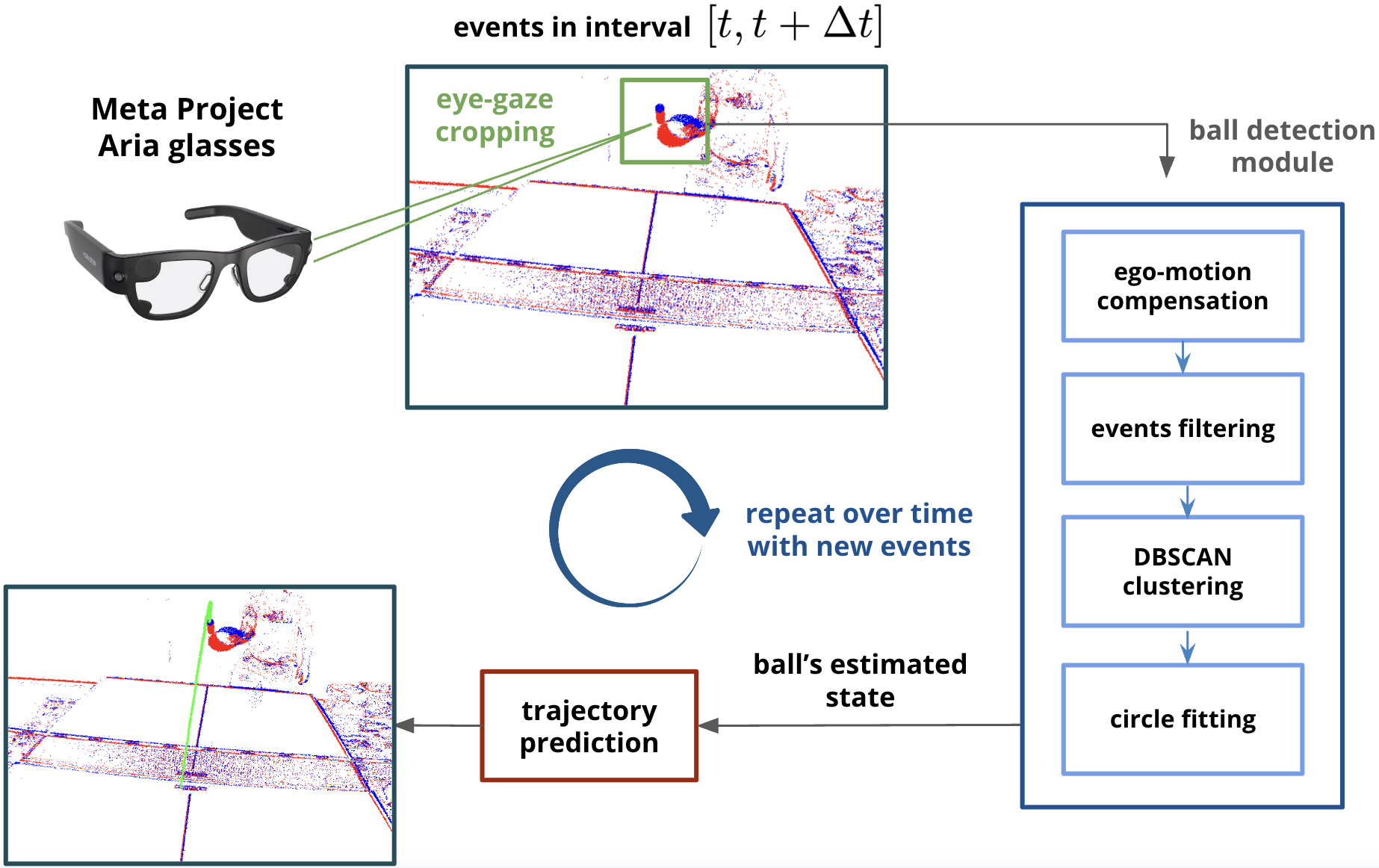}
     \caption{Overview of the method and its modules. The figure shows an iteration of the event-based pipeline, which processes events within a time window to predict a trajectory. 
}
     \label{img:method_overview}
\end{figure}

Our system relies on key modules, including ball detection, depth estimation, and trajectory prediction. On overview of the method is displayed in Figure~\ref{img:method_overview}. In the following sections, we describe each subsystem in detail.

\subsection{Ball Detection in the Event Domain}

\begin{figure*}[t]
    \centering
    \includegraphics[width=\textwidth]{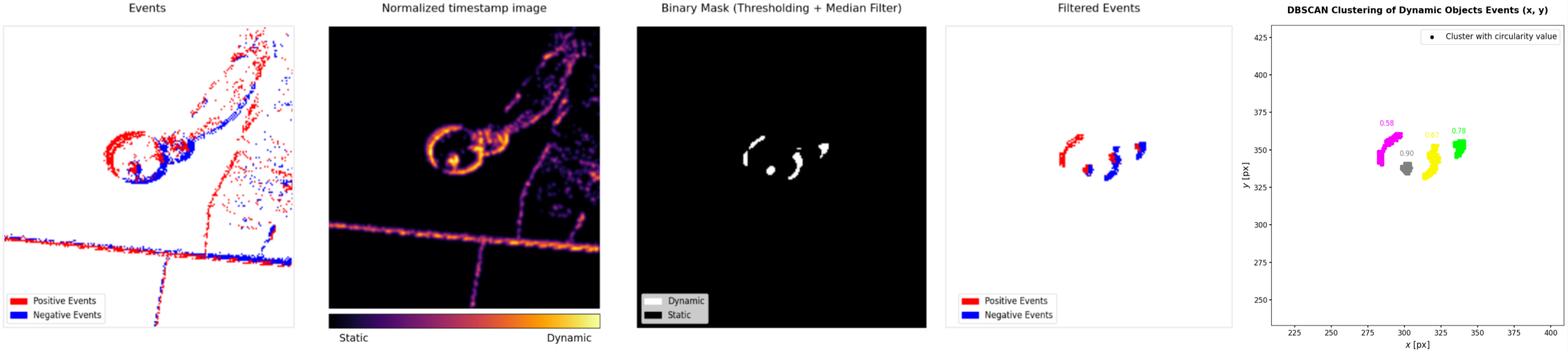}
    \caption{Pipeline of the ball detection stack. The leftmost image shows the input event visualization. The second image presents the normalized timestamp image obtained through motion compensation. The third image depicts the binary mask generated by applying the threshold $\theta_0 +\theta_1  w$, followed by a median filtering step. The forth image illustrates the final output events, highlighting the detected dynamic objects in the scene. The rightmost image shows the output of the DBScan filtering step on the normalized $(x, y, t)$ points, with the circularity value displayed on each cluster.}
    \label{fig:ball_detection_pipeline}
\end{figure*}

In our approach for detecting the ball in the event domain, we leverage information from the eye tracker embedded in the Meta’s Project Aria Glasses to reduce the bandwidth of the implementation. Specifically, we focus only on events that occur in a neighborhood of the eye-gaze's reprojection point on the image plane. Let \( \mathbf{x}_\text{ET} = (x_\text{ET}, y_\text{ET}) \) represent the coordinates of the eye-gaze reprojection on the event camera image plane, the events considered by the ball detection module belong to the subset:
\begin{equation}
    \mathcal{E} = \{ e_i \mid \mathbf{x}_i \in \mathcal{N}(\mathbf{x}_\text{ET}, w) \}_{i=0}^{N-1}
\end{equation}

where \( e_i = (\mathbf{x}_i, t_i, p_i)\) such that \( \{ t_i\}_{i=0}^{N-1} \in [t, t + \Delta t]\). $\mathcal{N}(\mathbf{x}_\text{ET}, w)$ represents a window of size $w \times w$ centered at $\mathbf{x}_\text{ET}$. By restricting the event set to this region, we not only process significantly fewer events, thus reducing bandwidth, but we also filter out potential outliers that could negatively impact the detection pipeline. Our approach assumes that the player's gaze is always directed toward the ball\footnote{This assumption was validated by monitoring the eye-gaze reprojection points of multiple players over different games, where it was observed that the players predominantly tracked the ball with their eyes}. In the early steps of our approach, we adopt a procedure similar to that of~\cite{falanga2020dynamic} and~\cite{forrai2023event}. Using the events in \( \mathcal{E} \), we apply motion compensation~\cite{8593805} \cite{falanga2020dynamic} to account for camera motion and remove static objects from the region of interest. The compensated coordinates \( \mathbf{x}_i^{mc} \) are defined as:
\begin{equation}
    \mathbf{x}_i^{mc} = K \left[ \mathbf{I} - [\bar{\omega}]_\times (t_i - t_0) \right] K^{-1} \mathbf{x}_i,
\end{equation}
where \( \mathbf{x}_i = (x_i, y_i) \), \( K \) is the intrinsic calibration matrix, \( \mathbf{I} \) is the identity matrix, \( [\bar{\omega}]_\times \) is the skew-symmetric matrix of the mean angular velocity \( \bar{\omega} \) obtained from the gyroscope measurements of the IMU mounted on the Aria, and \( t_0 \) is the reference time within the window. Following this, the motion-compensated mean timestamp image is computed as \( \mathcal{T}(\mathbf{x}) = \frac{\sum_i (t_i - t_0) \delta(\mathbf{x} - \mathbf{x}_i^{mc})}{\sum_i \delta(\mathbf{x} - \mathbf{x}_i^{mc})} \). To identify moving objects, we first compute the normalized timestamp image $\rho(\mathbf{x})$ and then generate a binary map \( B(\mathbf{x}) \) using an adaptive threshold, such that $ B(\mathbf{x})=1$ if $\rho(\mathbf{x}) > \theta_0 + \theta_1 \|\bar{\omega}\|$, while $B(\mathbf{x})=0$ otherwise. In this context, $\theta_0$ and $\theta_1$ are tuning parameters, $\theta_0$ determines the threshold level when the camera is not moving, whereas $\theta_1$ increases the threshold as the angular velocity grows. This binary map is used to filter out static objects, retaining events linked to dynamic ones. The remaining events are clustered using the DBSCAN algorithm, where we use the values \((t_i, x_i, y_i) \in \mathcal{E}_{dyn}\) as features for clustering the events, where \( \mathcal{E}_{dyn} = \{ e_i \mid B(\mathbf{x}_i )=1 \}_{i=0}^{N-1} \). The $x$ and $y$ components are normalized by dividing them by the image sensor's width and height, respectively. Meanwhile, time is scaled using min-max normalization within the chosen time window. Let \( S \) be the set of 2D points partitioned into clusters \( S_j \), obtained by collapsing the temporal dimention of the selected events. For each cluster, we compute its convex hull $\text{conv}(S_j)$ and evaluate its circularity:
\begin{equation}
    \gamma_j = \frac{{\text{P}(\text{conv}(S_j))}^2}{4\pi \cdot \text{A}(\text{conv}(S_j))}
\end{equation}
which should be close to $1$ for cluster of points having circular shape. The convex hull with the highest circularity value is selected, provided that its perimeter and area satisfy the following requirements \( \text{P}(\text{conv}(S_j)) \in [ \text{P}_{min}, \text{P}_{max}] \) and \( \text{A}(\text{conv}(S_j)) \in [ \text{A}_{min}, \text{A}_{max}] \), defined by the geometry of the problem. The cluster $\gamma^*$ meeting these criteria is identified as the ball. A visual overview of the method is shown in Figure~\ref{fig:ball_detection_pipeline}. Based on the the collected dataset (see Sect.~\ref{sec:results}), a $\Delta t = 5$ ms time window was experimentally validated to be sufficient for detecting both the fastest and slowest ball hits. Within these velocity ranges, enough events were generated, and their projections on the image plane remained well-approximated by a circumference. %

\subsection{Depth Estimation through Circle Fitting}

Accurately estimating the radius of a ping-pong ball at the opponent's distance is crucial for precise position and velocity estimation. 
We define \( \mathcal{E}_{\text{ball}} \) as the set of events associated with the detected ball, as a result of the motion compensation and DBSCAN filtering steps. Let \( \{ (t_i, \mathbf{x}_i') \}_{i=0}^{N_{{\text{ball}}}-1} \) be the set of image points belonging to \( \mathcal{E}_{\text{ball}} \), undistorted using known intrinsics $K_{ev}$ and distortion parameters $D_{ev}$ of the event camera, where each \( t_i \) represents a timestamp in the time interval \([0, T]\). We want to temporally divide this set into \(M\) equal batches defined as follows:
\begin{equation}
    \mathcal{B}_m = \left\{ (t_i, \mathbf{x}_i') \in \mathcal{E}_{\text{ball}}\mid \frac{(m-1)T}{M} \leq t_i < \frac{mT}{M} \right\}
\end{equation}
and $t_{\mathcal{B}_m} = \frac{(2m-1)T}{2M}, \; \text{for } m = 1, 2, \dots, M$, being the timestamp of the measurement inferred from $\mathcal{B}_m$, which is set to the midpoint of the interval. For the last batch, we include the endpoint \( T \) explicitly to ensure all points are assigned. For each $\mathcal{B}_m$, we select three points on \( \text{conv}(\mathcal{B}_m) \), denoted as \( \left. P_h  = (x_h, y_h) \right|_{\text{h}=1}^3  \), such that they maximize the sum of their pairwise Euclidean distances.
A circle is then fitted to these points, using the general equation of a circle:
\begin{equation}\label{eq:circle_equation}
    (x - \hat{x}_{m,\text{ball}})^2 + (y - \hat{y}_{m,\text{ball}})^2 = \hat{r}^2_{m,\text{ball}}
\end{equation} 
with \((\hat{x}_{m,\text{ball}}, \hat{y}_{m,\text{ball}})\) as the circle's center and \(\hat{r}_{m,\text{ball}}\) as its radius. Solving for the center and radius involves first solving the linear system derived from the three points, used to compute the center, and then obtaining the radius from Eq.~\ref{eq:circle_equation}. Once the radius is estimated, the depth \(Z_{m,\text{ball}}\) of the ball is calculated using the formula:
\begin{equation}
    \hat{Z}_{m,\text{ball}} = f \frac{W_{\text{metric}}}{\hat{r}_{m,\text{ball}}},
\end{equation}
where \(f\) represents the focal length of the camera, \(W_{\text{metric}} = 0.02\,\text{m}\) is the physical radius of the ping pong ball, and \(\hat{r}_{m,\text{ball}}\) is the estimated radius from the image. The circle fitting step ensures that \(\hat{r}_{m,\text{ball}}\) accurately reflects the image-space radius, minimizing errors in depth estimation.

\label{sec:ball_dection_subsection}

\subsection{Trajectory Estimation using Ball Dynamics}
The output of the ball detection pipeline is a set of measurements \( \{ (t_{\mathcal{B}_m}, \hat{x}_{m,\text{ball}}, \hat{y}_{m,\text{ball}}, \hat{Z}_{m,\text{ball}}) \}_{m=0}^{M-1} \), representing the extracted ball's center image coordinates and depth at the computed timestamps. From this point onward, we will simplify our notation by using $\hat{x}_m$ instead of $\hat{x}_{m,\text{ball}}$. \newline

\textbf{Monotonically Constrained Polynomial Regression}: Given $M$ data points, we aim to fit a polynomial regression model of degree \(d\) to each measurement variable independently, namely $x(t)$, $y(t)$ and $Z(t)$. The functions are estimated by minimizing the least squares error between the observed data points and the fitted polynomial model. However, we add a condition to the estimation of $Z(t)$, which we constrain to be monotonically decreasing. The motivation behind this constraint is that the ball’s trajectory is always directed toward the camera, implying a consistently decreasing depth, since we focus exclusively on sequences that begin precisely at the moment the opponent strikes the ball. By enforcing this condition, we effectively filter out potential outlier measurements in the circle fitting module that do not conform to this expected motion pattern. Function $Z(t)$ is therefore estimated by solving the following optimization problem:
\begin{equation}
        \min_{\beta_Z}  \frac{1}{M}\sum_{m=1}^{M} ( \hat{Z}_m - \sum_{j=0}^{d} \beta_{Z,j} t_{{\mathcal{B}_m}}^j )^2
        \quad \text{s.t.} \quad \dot{Z}(t)  \leq 0
\end{equation}

with \(\beta_{Z,j}\) being the polynomial coefficients of $Z(t)$. The estimation of $x(t)$ and $y(t)$ follows the same formulation but without adding the constraint on the derivative. From these estimations, we then reconstruct the positions $\mathbf{p}_k$ of the ping-pong ball in the 3D space, and get the trajectory data $\{ t_{{\mathcal{B}}_m}, \hat{\mathbf{p}}_m, \hat{\mathbf{v}}_m \}_{m=1...M}$, which serve as the initial knowledge of the system’s state within the chosen interval $T$. The estimated 3D position of the ball is therefore computed as \( \hat{\mathbf{p}}_m = Z(t_{{\mathcal{B}_m}}) \cdot K_{ev}^{-1} \cdot \mathbf{x}(t_{{\mathcal{B}_m}}) \), while the velocity \( \hat{\mathbf{v}}_m = (\hat{v}_{x, m}, \hat{v}_{y, m}, \hat{v}_{z, m}) \) is computed using finite differences. The number $M$ of samples is a design parameter set at runtime, and this step is always applied at every iteration. %

\vspace{2mm}
\textbf{Physics-based Differential Equations with Extended Kalman Filter bootstrapping}: To propagate the trajectory of a flying ball into the future using the dynamical system’s differential equations, it is critical to estimate an accurate initial position \( \mathbf{p}_0 \) and velocity \( \mathbf{v}_0 \). One way is to set the initial position to $\mathbf{p}_0 = \hat{\mathbf{p}}_0$ and the initial velocity to the average velocity over the measurements $ \mathbf{v}_0 = \frac{1}{M}\sum_{m} \hat{\mathbf{v}}_m$. By rewriting \( \mathbf{p}(t) = \mathbf{p_0} + \int_0^t \mathbf{v}(t') \, dt' \) and  \( \mathbf{v}(t) = \mathbf{v_0} + \int_0^t \mathbf{a}(t') \, dt' \) into their discrete forms while accounting for gravitational acceleration $F_g$ and the drag force $F_d$, we estimate the future trajectory by iteratively updating the following: 
\begin{equation}
    \mathbf{p}(t_i) = \mathbf{p}(t_{i-1}) + \mathbf{v}(t_{i-1}) \Delta t 
\end{equation}
\begin{equation}
\mathbf{v}(t_i) = \mathbf{v}(t_{i-1})  -k_d |\mathbf{v}(t_{i-1})|
\mathbf{v}(t_{i-1}) \Delta t + g \, \Delta t
\end{equation}

where $\Delta t = t_i - t_{i-1}$. To further improve the initial conditions, we extend the previous method by introducing an Extended Kalman Filter formulation, which models the motion of the ping-pong ball under the assumption of constant acceleration, with the state vector being defined as $\mathbf{x}^{ekf} = [\mathbf{p}, \mathbf{v}, \mathbf{a}]^T$, and the state transition and measurement model Jacobians being approximated as:
\begin{equation}
\mathbf{F} =
\begin{bmatrix}
\mathbf{I}_3 & \Delta t \; \mathbf{I}_3 & \frac{1}{2} \Delta t^2 \; \mathbf{I}_3 \\
\mathbf{0}_3 & \mathbf{I}_3 & \Delta t \; \mathbf{I}_3 \\
\mathbf{0}_3 & \mathbf{0}_3 & \mathbf{I}_3
\end{bmatrix}, \quad \mathbf{H} = \left[ \mathbf{I}_6 \quad \mathbf{0}_{6 \times 3}\right]
\end{equation}

Specifically, we initialize the EKF with $\mathbf{p}_0 = \hat{\mathbf{p}}_0$, $\mathbf{v}_0 = \hat{\mathbf{v}}_0$, and $\mathbf{a}_0 = \frac{F_g + F_d}{\textbf{m}}$, with $\textbf{m}$ being the mass of the ball. The EKF then undergoes $M$ predict-update iterations, progressively refining its state estimates. After these iterations, the final estimated position and velocity, denoted as $\mathbf{p}_M^{ekf}$ and $\mathbf{v}_M^{ekf}$, serve as the initial conditions for the dynamical system’s differential equation approach. That is, instead of setting $\mathbf{p}_0$ and $\mathbf{v}_0$ directly from raw measurements, we use $\mathbf{p}_0 = \mathbf{p}_M^{ekf}$ and $\mathbf{v}_0 = \mathbf{v}_M^{ekf}$. This hybrid approach allows for a more robust trajectory prediction by leveraging both statistical filtering and physical modeling. %

%% file: sec/4_latency_analysis.tex
\section{Latency Analysis} \label{sec:latency_analysis}
For a vision system to achieve real-time perception comparable to human visual latency, it should ideally process images within 10–50 ms~\cite{pulli2012real}. The human visual system has inherent delays: basic light perception occurs in approximately 13 ms, motion perception takes around 30–60 ms~\cite{moutoussis1997direct}, and full scene understanding requires 80–100 ms~\cite{thorpe1996speed}. To match human reaction times, image processing should aim for a latency of under 50 ms per frame. However, standard RGB cameras introduce delays due to exposure times ranging from 1 to 100 ms, leading to motion blur and slower processing time. As already mentioned before, event cameras enable significantly lower latency, capturing high-speed changes in brightness asynchronously.

\textbf{Computational latency}: In this context, previous work~\cite{falanga2020dynamic} demonstrated a low-latency event-based dynamic obstacle detection system for an autonomous quadrotor, achieving a computational latency of $3.56$ ms by measuring processing time of the events collected in a $10$ ms time window. This latency represents the time from when events are received until the first avoidance command is issued, ensuring timely obstacle detection. Although our detection pipeline follows a similar structure to that of~\cite{falanga2020dynamic} and \cite{forrai2023event}, it achieves lower latency due to key implementation changes. Since the obstacles we detect are smaller and typically located at greater distances, fewer events are required to represent them. Additionally, we leverage eye gaze reprojection to crop a region of interest, discarding irrelevant events and reducing computational overhead. A detailed breakdown of the computation times of the pipeline is provided in Table~\ref{tab:latency_breakdown}. While still considering $10$ ms time windows, our method achieves an overall mean computational latency of $2.35$ ms across all collected sequences, which is lower than that of~\cite{falanga2020dynamic}. Following the previous latency analysis, Table~\ref{tab:latency_comparison} presents the latency evaluation for the cropping of the region of interest (ROI) around the eye-gaze vector reprojection. In this case, we consider $5$ ms time windows, corresponding to the time intervals for each ball detection. We compare the mean computational latency with and without cropping, using data from all collected trajectories. The results highlight a substantial reduction in latency when applying cropping, with the average processing time dropping from $16.18$ ms to just $1.5$ ms. Additionally, the average number of events processed per trajectory decreases significantly, from $7157$ to $735$, demonstrating the efficiency of this approach. All evaluations were conducted on a CPU-only setup using an Intel Core i7-13700H (14 cores, 20 threads), by setting $w = 80$ pixels.

\begin{table}[ht]
    \centering
    \begin{tabular}{lccc}
        \toprule
        \textbf{Step} & $\mu$ [ms] & $\sigma$ [ms] & \textbf{Perc. [\%]} \\
        \midrule
        Ego-Motion Comp. & 1.390 & 0.775 & 59.1 \\
        Thresh. + Morph. Ops. & 0.083 & 0.025 & 3.5 \\
        DBSCAN + filtering & 0.879 & 0.455 & 37.4 \\
        \midrule
        \textbf{Total} & 2.352 & 1.146 & 100 \\
        \bottomrule
    \end{tabular}
    \caption{Computational time distribution for each step of the proposed pipeline, including mean, standard deviation, and percentage breakdown of the total execution time.}
    \label{tab:latency_breakdown}
\end{table}

\begin{table}[ht]
    \centering
    \begin{tabular}{l c c c}
        \toprule
        \textbf{Method} & $\mu$ [ms] & $\sigma$ [ms] & \textbf{\# Events}\\
        \midrule
        w/ ROI cropping  & 1.497 & 0.696 & 735\\
        w/o ROI cropping & 16.183 & 6.807 & 7157\\
        \bottomrule
    \end{tabular}
    \caption{Comparison of the overall latency of the method and the average number of events (averaged over the entire dataset) being processed, with and without region of interest cropping around the eye-gaze reprojection.}
    \label{tab:latency_comparison}
\end{table}

%% file: sec/5_experiments.tex
\section{Results} \label{sec:results}
In this section, we describe the experimental setup and methodology used to validate our approach. We first outline the hardware configuration and data collection process, followed by the evaluation procedure. Finally, we present the results and analysis of our findings.

\subsection{Hardware Setup and Data Collection}

Our experimental setup relies on multiple sensors for comprehensive data collection and accurate validation. The primary recording device is the \textit{Meta Project Aria} glasses. We use recording profile $\#28$, which captures RGB images at $30$ FPS with a resolution of $1408 \times 1408$, eye-tracking data at $60$ FPS, audio recordings from seven microphones at $48$ kHz, SLAM data including point cloud map of the surrounding and pose of the glasses at $30$ FPS and measurements from the two available IMU sensors on the glasses at $1$ kHz and $800$ Hz respectively. Additionally, we incorporate an \textit{iniVation DVXplorer} event camera with a 6 mm focal length and $640 \times 480$ resolution (VGA). This camera includes an IMU operating at $800$ Hz. To obtain ground-truth trajectory data, we use an \textit{OptiTrack} motion capture system, which records the 3D trajectory of the ping pong ball at $200$ Hz and the 6D pose of the Aria glasses with respect to a world reference frame. A significant part of the project involved collecting extensive multi-sensor data for validation. In total, we recorded 30 gaming sequences with five participants, ensuring diversity in gameplay conditions. To enhance eye-tracking accuracy, each recording session began with individualized eye gaze calibration. 

\textbf{Data Synchronization and Calibration:} To achieve precise data synchronization, IMU gyroscope readings from the Aria glasses and event camera were aligned in the frequency domain to identify peak correlations with millisecond accuracy~\cite{Furrer2017FSR}. Additionally, synchronization with OptiTrack was facilitated by detecting audio peaks from Aria microphones corresponding to ping pong ball bounces, with a default bounce at the start of each session serving as a temporal reference. Calibration involved stereo calibration to determine the intrinsic parameters and relative transformation between the Aria RGB and event cameras, using Kalibr toolbox~\cite{6696514}, along with hand-eye calibration~\cite{Furrer2017FSR} to align the Aria RGB camera with the ground-truth pose recorded by OptiTrack, achieved by attaching markers to the glasses.

\subsection{Performance evaluation}

In this section, we present a comprehensive quantitative evaluation of the performance of our algorithm in detecting and predicting the trajectory of a ping-pong ball using event cameras. We validate our method by measuring two metrics: one for the ball detection algorithm and another one to evaluate the trajectory prediction method. 

\textbf{Detection Success Rate Analysis:} To evaluate the success rate of our detection algorithm, we measure the 2D error norm between the detected ball's center and the reprojection of the 3D ground truth position at the corresponding timestamp in the event image space. A detection is considered successful if the error is smaller than a tolerance value $\epsilon$ that we set to $5$ pixels in our evaluation.

\begin{table}[ht]
    \centering
    \begin{tabular}{lcc}
        \toprule
        \textbf{$\theta_1$} & \textbf{Detection Rate (\%)} & \textbf{\# Events} \\
        \midrule
        0.6 & 88.88 & 1009 \\
        0.8 & 92.59 & 701 \\
        1.4 & 59.26 & 279 \\
        \bottomrule
    \end{tabular}
    \caption{Detection rates and average number of events after motion compensation and thresholding, for different values of $\theta_1$.}
    \label{tab:detection_success_rate}
\end{table}

Table~\ref{tab:detection_success_rate} presents detection rates and the average number of events representing the ball for different values of \( \theta_1 \), while keeping  \( \theta_0 \) fixed, obtained by running the ball detection algorithm on the recorded game sequences. From the results, we determine that setting \( \theta_1 = 0.8 \) yields the highest success rate. Increasing the threshold beyond this value leads to a decrease in detection performance. As the threshold increases, more events associated with slower-moving balls are filtered out, reducing their detectability, even if that corresponds to a decreased bandwidth usage due to less events being processed. On the other hand, a very low threshold may fail to adequately filter static objects, leading to potential false positives. In addition to the previous findings, Table~\ref{tab:detection_success_rate_cropping_ROI} examines the impact of applying cropping around eye-gaze. The results indicate that cropping not only significantly reduces bandwidth, as previously demonstrated in Table~\ref{tab:latency_comparison}, but also enhances detection performance. This improvement occurs because focusing on the area surrounding the ball helps eliminate outliers and other circular objects in the scene that could be mistakenly identified as the ball if they were not properly filtered out earlier.

\begin{table}[ht]
    \centering
    \begin{tabular}{l c}
        \toprule
        \textbf{Method} & \textbf{Detection Rate (\%)} \\
        \midrule
        w/ ROI cropping    & 92.59  \\
        w/o ROI cropping & 81.48  \\
        \bottomrule
    \end{tabular}
    \caption{Comparison of the detection rate of the method with and without cropping around the eye-gaze reprojection. }
    \label{tab:detection_success_rate_cropping_ROI}
\end{table}

\begin{table*}[ht]
    \centering
    \begin{tabular}{ccccc}
        \toprule
        Method & Update freq. (Hz) & $T_{pred} = 10$ ms & $T_{pred} = 20$ ms & $T_{pred} = 33$ ms \\
        \midrule
        \midrule
        Diff. Eq. & 200 & $0.784 \pm 0.478$ & $0.382 \pm 0.288$ & $\textbf{0.242} \pm \textbf{0.096}$ \\
        Diff. Eq. & 30 & - & - & $0.409 \pm 0.504$ \\
        DCGN~\cite{8957482} & - & $0.741 \pm 0.356$ & $0.514 \pm 0.270$ & $0.442 \pm 0.194$ \\
        \rowcolor{lightgray} 
        \midrule
        $\text{Diff. Eq.}^\star$ & 200 & $0.215 \pm 0.321$ & $0.083 \pm 0.06$ & $0.052 \pm 0.025$ \\
        \bottomrule
    \end{tabular}
    \caption{A comparison of the performance of different trajectory prediction modalities using single-batch forecasting approach. The table presents the Root Mean Squared Error (RMSE $\downarrow$) in meters of the predicted impact point across all collected trajectories. ${}^\star$The physics-based differential equation was fitted with ground truth states instead of the measurement from the detection pipeline.}
    \label{tab:single-batch-rmse-eval}
\end{table*}

\textbf{Trajectory Prediction Performance:} We assess the performance of the trajectory prediction pipeline by analyzing impact point accuracy by calculating the squared error of the impact points on the table over all the collected trajectories. The results of this experiment are presented in Table~\ref{tab:single-batch-rmse-eval} and Table~\ref{tab:online-rmse-eval}, differing from the way we deploy the trajectory prediction pipeline. In one approach a single time window of events is used to predict the future trajectory in a single-batch estimation, which is then propagated over time using relative poses from the Aria. This method is computationally efficient since it requires running the method only once, but its accuracy is limited due to reliance on a restricted amount of data. Alternatively, the predicted trajectory can be recomputed at each new ball measurement, taking into account all past measurements. A visual representation of the outcome of the trajectory prediction pipeline is presented in Figure~\ref{fig:prediction_sequences}.

\textit{a) Single-Batch Forecasting Performance:} We analyze how trajectory prediction performance changes as we increase the time window used to accumulate event data, with each measurement being obtained from a $5$ ms time window. For this investigation, we do not use EKF bootstrapping because of the short prediction time windows taken into account, which result in comparable results. To highlight the advantage of event cameras over frame-based ones, we compare against a differential equation prediction model whose initial conditions are set from low-frame-rate raw measurements. This baseline follows the same pipeline as our method but only receives data at conventional frame-based camera intervals. For Project Aria Glasses, the maximum frame rate is $30$ FPS, meaning measurements can occur at least every $33$ ms. Additionally, we compare our approach to~\cite{8957482}, trained on ground truth trajectories upsampled to $800$ Hz. The results in Table~\ref{tab:single-batch-rmse-eval} show that setting the initial conditions from high-frame-rate measurements yields the lowest error. It is worth mentioning that non-negligible error occur even when using ground truth to predict the future trajectory, setting a lower bound on achievable performance using such short time prediction intervals. Overall, while performance is weak with $T_{pred}=10$ ms and $T_{pred}=20$ ms, it improves significantly when using $T_{pred}=33$ ms. In contrast, the differential equations prediction model initialized with Project Aria frame-rate updates exhibits much higher error due to its reliance on only two measurements, which, as discussed in Section~\ref{sec:method}, can be noisy. However, the model performing the worst in this evaluation is DCGN~\cite{8957482}, showing the highest error even with $T_{pred}=33$ ms. This suggests that traditional dynamical model fitting is more effective than other methods when prediction time windows are short.

\textit{b) Online Forecasting Performance:} We analyze the impact of continuously recomputing the ball’s future trajectory as new measurements become available in our pipeline. Specifically, we define a time horizon of $0.2$ seconds within which trajectory updates are allowed. Table~\ref{tab:online-rmse-eval} presents a comparison of different baseline methods, evaluated on the most recent predicted trajectory that incorporates the highest number of measurements. We evaluate a standard differential equation-based approach initialized directly from raw measurements, the EKF-bootstrapping variant, the low-update-rate version at $30$ Hz, and the learning-based DCGN method~\cite{8957482}. Overall, we observe a significant improvement in performance compared to the results in Table~\ref{tab:single-batch-rmse-eval}. Among the tested methods, DCGN achieves the best accuracy, with an RMSE of $0.1072$ m. The next best approach is the differential equation model with EKF bootstrapping, which outperforms the version without bootstrapping. This confirms that EKF bootstrapping provides a more reliable initialization, leading to slightly improved accuracy. On the other hand, the low-frame-rate differential equation method exhibits the worst performance, highlighting the importance of high-frequency updates to effectively filter out outliers.

\begin{table}[ht]
    \centering
    \begin{tabular}{ccc}
        \toprule
        Method &  Update freq. (Hz) & $ \left. \textit{E}(t) \; \right|_{\;t = T_{pred}^{MAX}} \, \text{[m]}$ \\
        \midrule
        \midrule
        Diff. Eq.   & 200  & $0.1472 \pm \textbf{0.0938}$  \\
        Diff. Eq. (EKF) & 200  & $ 0.1432 \pm 0.0991 $ \\
        Diff. Eq.  & 30  & $0.1915 \pm 0.1705$  \\
        DCGN~\cite{8957482} & -  & $\textbf{0.1072} \pm 0.1085$ \\
        \midrule
        \rowcolor{lightgray} 
        $\text{Diff. Eq.}^\star$ & 200 &  $ 0.0121 \pm 0.0075$\\
        \bottomrule
    \end{tabular}
    \caption{The table shows the performance using the online forecasting approach, comparing the RMSE between the predicted impact point of the last computed trajectory within a fixed prediction horizon of $0.2$ s and the ground truth, averaged over the dataset.}
    \label{tab:online-rmse-eval}
\end{table}

%% file: sec/6_discussion.tex
\section{Discussion} \label{sec:discussion}
The current work demonstrates the feasibility of real-time ping pong ball trajectory prediction from the player's view. Nonetheless, our implementation has some limitations, primarily due to hardware constraints and the challenges of an egocentric setup. First of all, the egocentric perspective introduces complexities not present in static or externally mounted systems. The player's head movements trigger events across the entire scene, making it difficult to isolate those corresponding to the ball, especially when it blends into the background with other moving objects, such as an opponent. Camera placement further complicates trajectory prediction, as the small ball must be detected at a relatively far distance compared to its size, introducing noisier measurements and lower precision. One way to mitigate this, would be relying on a higher-resolution camera.
Although using eye-gaze tracking significantly improves efficiency by reducing bandwidth, it also introduces a dependency on human behavior, which can be unpredictable. If the user briefly looks away, detection and tracking may fail, reducing robustness. Expanding the cropping window around the eye-gaze image reprojection can mitigate this issue but at the cost of increased bandwidth usage and reduced efficiency. Another limitation is the absence of an automatic trigger to detect the moment the opponent hits the ball during continuous gameplay. Future work could address this by training a neural network to automatically infer it from audio signals and trigger the perception pipeline accordingly.

%% file: sec/7_conclusions.tex
\section{Conclusions}
We introduce the first real-time, event-based perception system for table tennis trajectory prediction using Meta Project Aria glasses in a monocular, egocentric setup. While not yet integrated into a full AR/VR application, our work demonstrates the feasibility of this approach and paves the way for future real-time sports analysis from an egocentric vision perspective, even though further improvements are needed. Our method showcases the benefits of event-based perception for low-latency tasks, effectively overcoming the bandwidth-latency trade-off of traditional cameras. The event camera’s high temporal resolution enables more frequent measurements, making the system more robust to outliers and leading to more accurate predictions of the ball’s future position. Additionally, by leveraging eye-gaze tracking to focus on regions of interest and employing a lighter obstacle detection method, our system achieves lower latency compared to previous approaches.
\label{sec:conclusions}

%% file: sec/X_suppl.tex
\clearpage
\setcounter{page}{1}
\maketitlesupplementary

\section{The aerodynamics model of a ping-pong ball}

In this section, we provide an overview of the aerodynamics model of a ping pong ball used in the paper, focusing on the equations of motion that describe its trajectory. A standard ping-pong ball moving through the air experiences four primary forces: gravitational force ($F_g$), buoyancy force ($F_b$), drag force ($F_d$), and Magnus force ($F_m$). For our investigation, we can ignore the buoyancy force $F_b = -m_b g$, because the mass of the air displaced by the ping-pong ball is negligible with respect to the mass of the ball $m$. We also neglect the spin of the ball (Magnus force component $F_m$), cause it is not directly observed by the vision system due to the small dimension of the ball. Therefore, the sum of the forces acting on the ball can be expressed as \( \sum \mathbf{F} = \mathbf{F}_g + \mathbf{F}_d \). The gravitational force is given by $\mathbf{F_g} = -mg$, where $m$ represents the mass of the ball, and $g = [0, 0, -9.81]^T$ is the acceleration due to gravity. The drag force, which opposes motion through the air, follows the equation \( \mathbf{F_d} = - \frac{1}{2} C_d \rho A |\mathbf{v}(t)| \mathbf{v}(t) \), where \( C_d \) is the drag coefficient, \( \rho \) is the air density, \( A \) is the cross-sectional area of the ball and \( |\mathbf{v}(t)| \) is the magnitude of the velocity vector \( \mathbf{v}(t) \). By substituting these forces, we obtain:

\begin{equation}
\sum \mathbf{F} = mg - \frac{1}{2} C_d \rho A |\mathbf{v}(t)| \mathbf{v}(t)
\end{equation}

This simplifies the equation of motion to:

\begin{equation}
\dot{\mathbf{v}}_k(t) = g - k_d |\mathbf{v}(t)| \mathbf{v}(t)
\end{equation}

where we set $k_d = \frac{C_d \rho A}{2m}$. For a standard ping-pong ball, the known values are: $\rho= 1.225 kg/m^3$, $r = 0.02m$, $m = 0.0027kg$ and $C_d = 0.4$. We additionally model the motion of the ball by introducing a simplified bouncing model. When the estimated \( z \)-coordinate of the ball is lower than \( h_{\text{table}} \) (determined using ArUco marker detection, as shown in Fig.~\ref{fig:prediction_sequences}) and \( \mathbf{v}_z \) is negative, we switch to the bounce model:
\[
\mathbf{v}_z^+ = e \, \mathbf{v}_z^- , \quad \text{with } 0 < e < 1.
\]

Here, \( \mathbf{v}_z^- \) represents the velocity component along the vertical axis just before impact, and \( \mathbf{v}_z^+ \) is the velocity just after. This model accounts for energy loss upon impact due to inelastic collisions with the table.

\subsection{Including Rotational Dynamics}

A more precise representation of the ball’s motion should account for its rotational dynamics, particularly the Magnus force, $F_m$, which influences the ball’s trajectory. This force arises due to the interaction between the ball’s spin and the surrounding air, significantly affecting its movement, and it is defined as follows:

\begin{equation}
    F_m =  C_m \rho  A r (\omega \times v)
\end{equation}

where $C_m$ is the Magnus coefficient and $\omega$ is the angular velocity of the ping-pong ball. The equation of motion including the Magnus force would then become:

\begin{equation}
\dot{\mathbf{v}}_k(t) = g - k_d \|\mathbf{v}(t)\| \mathbf{v}(t) + k_m (\omega \times v)
\end{equation}

or in its discrete formulation:

\begin{multline}
    \mathbf{v}(t_i) =  \mathbf{v}(t_{i-1}) +
    \begin{bmatrix}
        -k_d \|\mathbf{v}\| & -k_m \omega_z & k_m \omega_y \\
        k_m \omega_z & -k_d \|\mathbf{v}\| & -k_m \omega_x \\
        -k_m \omega_y & k_m \omega_x & -k_d \|\mathbf{v}\|
    \end{bmatrix} \\  \mathbf{v}(t_{i-1}) \Delta t + \begin{bmatrix} 0 \\ 0 \\ -g \end{bmatrix} \Delta t
\end{multline}

where $k_m = \frac{ C_m \rho  A r}{m}$. When incorporating rotational dynamics into the motion model, setting the initial conditions of the differential equation requires also specifying an initial estimate of the ball's angular velocity. However, this quantity is not directly measurable from our observations, but it can be inferred from the trajectory data $\{ t_{{\mathcal{B}}_k}, \hat{\mathbf{p}}_k, \hat{\mathbf{v}}_k \}_{k=1...K}$ estimated from the measurements.

\section{Sensing Latency Analysis}
\begin{figure}[t]
     \vspace{2mm}
     \centering
     \includegraphics[width=0.49\textwidth]{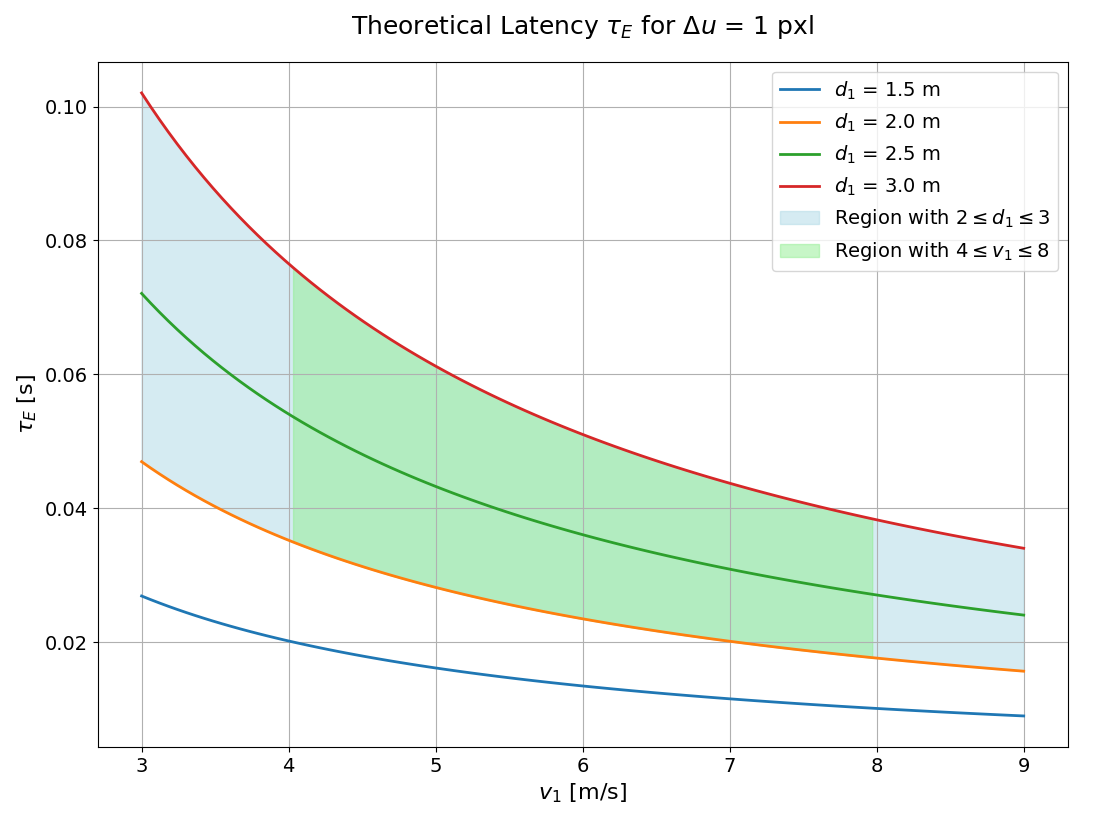}
     \caption{The sensing latency $\tau_E$ of an event camera with $640 \times 480$ resolution and a focal length of $6$ mm. The shaded green region represents the ideal sensing latency conditions based on our dataset, where the relative velocity between the ball and the Project Aria glasses varies from approximately $\sim4$ m/s to $\sim8$ m/s.}
     
     \label{fig:theoretical_latency_characterization}
\end{figure}

We present an analysis of the sensing latency of our algorithm, which refers to the time window required to detect motion events and produce reliable results. As described in \cite{falanga2019howfast}, an obstacle is detected using an event camera when its edges generate an event. This occurs when the relative motion between the camera and the obstacle causes a significant intensity change, triggering an event. Prior work \cite{falanga2019howfast} has shown that an obstacle's edge produces an event when its projection on the image plane moves by at least one pixel. As already shown in~\cite{falanga2019howfast}, the time required for an obstacle to traverse a pixel distance $\Delta u = 1$ in the image plane is given by:

\begin{equation}
\tau_E = \frac{1}{\hat{\mathbf{v}}} \frac{\Delta u d^2}{f r_o + \Delta u d}
\end{equation}

where $\hat{\mathbf{v}}$ is the object's relative velocity with respect to the camera, $d$ represents the obstacle’s distance along the camera’s optical axis, $r_o$ is the obstacle’s radius, and $f$ is the camera’s focal length. This calculation assumes that the optical axis passes through the geometric center of the obstacle, which is approximated as a segment. Figure~\ref{fig:theoretical_latency_characterization} illustrates the theoretical sensing latency fo an event camera to perceive a $\Delta u = 1$ pixel motion in the image plane of a ping-pong ball, as a function of distance $d$ and speed $\hat{v}$. In our specific case, we aim to track the ball when struck by the opponent’s racket, therefore the ball is typically observed at distances $d$ ranging from $2$ to $3$ meters.

\section{Deep Conditional Generative Network}

\begin{figure*}[t]
    \centering
    \subfloat[ ]{\includegraphics[width=0.34\textwidth]{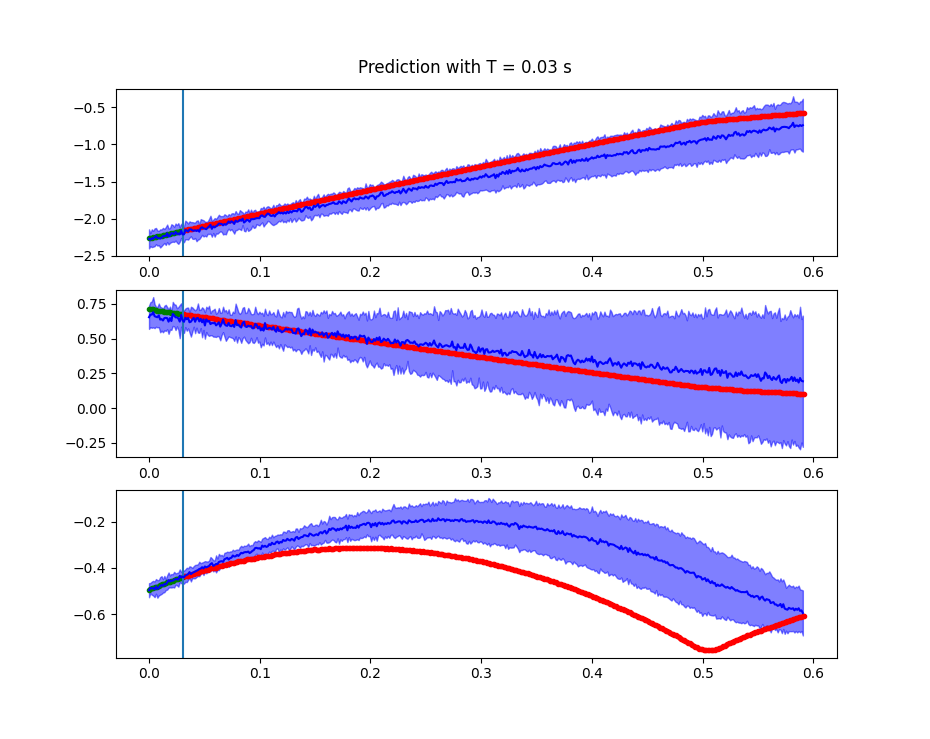}}
    \subfloat[ ]{\includegraphics[width=0.34\textwidth]{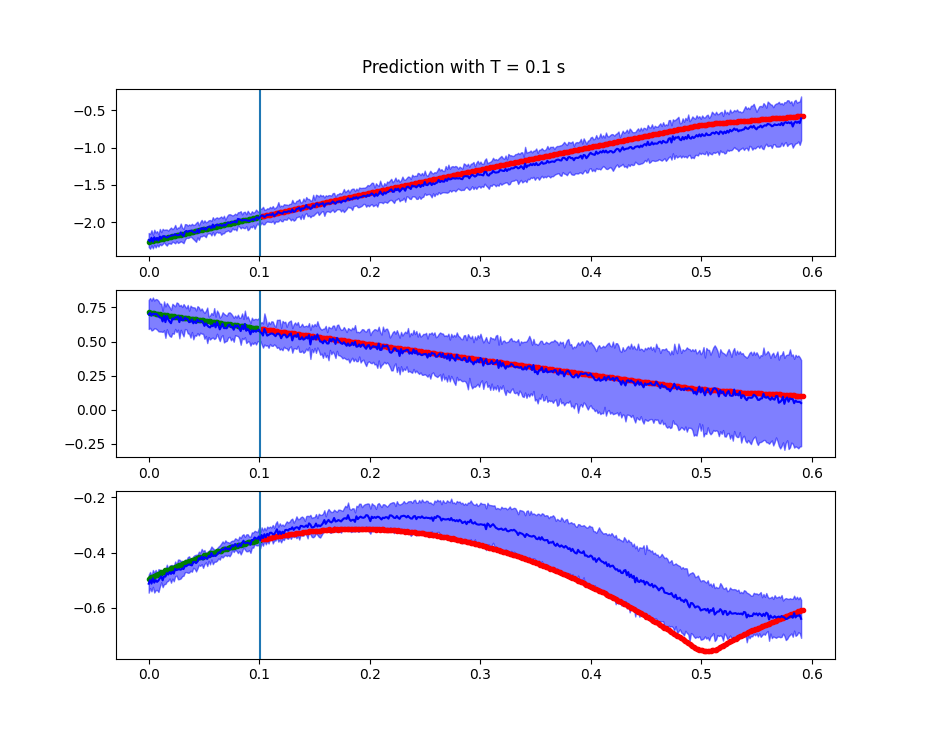}}
    \subfloat[ ]{\includegraphics[width=0.34\textwidth]{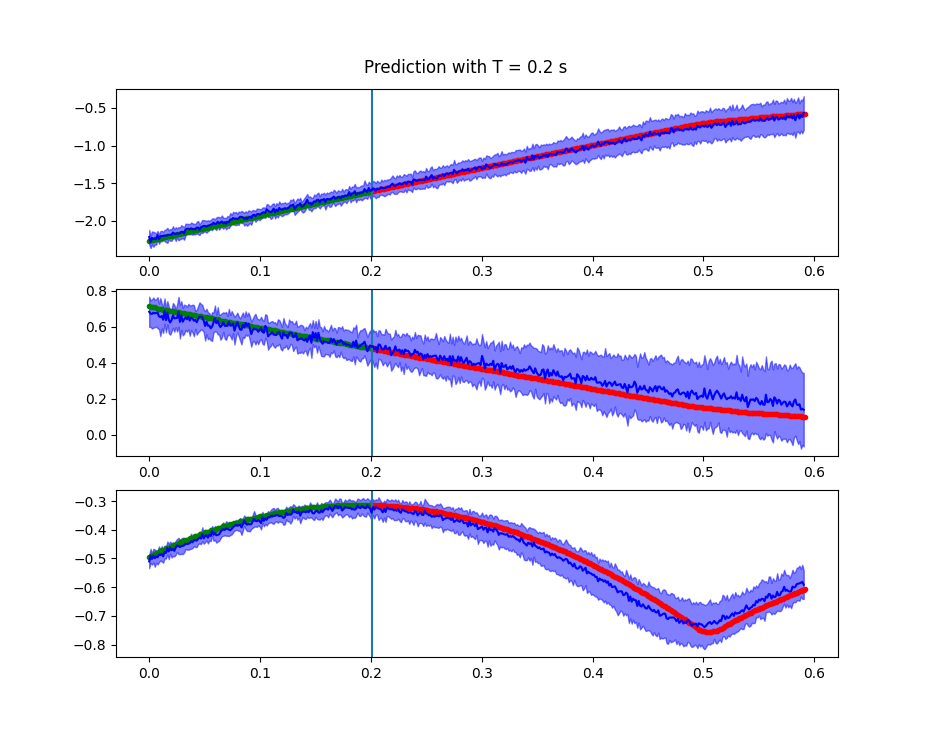}}
    \caption{Visualization of the predicted trajectory compared to the ground truth for different prediction time horizon values (a) \( T = 0.03 \) s, (b) \( T = 0.1 \) s, and (c) \( T = 0.2 \) s. The x, y, and z components of the trajectory are shown, where the green segments represent the input to the network (before the split), the red segments represent the ground truth after the split, and the blue lines indicate the predicted trajectory with the $3 \sigma$ standard deviation.}
    \label{fig:trajectory_predictions}
\end{figure*}

An overview of the variational autoencoder network introduced in~\cite{8957482}, which is employed for trajectory prediction, and the outcomes of which have been discussed in Section~\ref{sec:results}, is provided in this paragraph. The work proposes a Deep Conditional Generative Network (DCGN) for real-time trajectory prediction, mapping partial trajectories to a latent Gaussian space to predict future points. In this framework, a trajectory is represented as a probability distribution conditioned on observed data. Let $\mathbf{x}_{1:t}$ denote the observed trajectory up to time $t$, and $\mathbf{x}_{t+1:T} $ represent the future trajectory to be predicted. The model efficiently learns the conditional distribution $p(\mathbf{x}_{t+1:T} | \mathbf{x}_{1:t}) = \int p(\mathbf{x}_{t+1:T} | \mathbf{z}, \mathbf{x}_{1:t}) \, p(\mathbf{z} | \mathbf{x}_{1:t}) \, d\mathbf{z}$, which is achieved by introducing a latent variable $\mathbf{z}$ that captures the underlying dynamics of the trajectory. The training procedure involves maximizing the evidence lower bound (ELBO) to approximate the true posterior distribution. This method enables better long-term prediction of complex trajectories compared to LSTMs and differential equations, thanks to its probabilistic modeling, uncertainty estimation, and efficient latent space representation.

To assess the predictive capabilities of the DCGN model, we conducted an additional evaluation on the ground truth trajectories. By using different prediction horizon lengths \( T \), we analyzed how well the model can forecast future motion while minimizing error. The obtained results, presented in Table~\ref{tab:rmse_results}, show that larger prediction horizons generally lead to better performance, as indicated by lower Root Mean Squared Error (RMSE) values.

\begin{table}[ht]
    \centering
    \begin{tabular}{cc}
        \toprule
        \textbf{Horizon Time} & \textbf{RMSE} \\
        \midrule
        0.01 & 0.2616 \\
        0.03 & 0.2055 \\
        0.12 & 0.1737 \\
        0.20 & 0.1470 \\
        0.30 & $\textbf{0.1177}$ \\
        \bottomrule
    \end{tabular}
    \caption{RMSE values of the predicted trajectory across the entire dataset for different horizon prediction times \( T \).}
    \label{tab:rmse_results}
\end{table}

The DCGN model was trained on ground truth trajectories upsampled to 0.8 kHz, using an 80/20 split for training and validation. We observed that for short horizons, the model struggles to produce accurate predictions, resulting in high RMSE values. Figure~\ref{fig:trajectory_predictions} visually compares the predicted and ground truth trajectories for three selected horizon values: \( T = 0.03 \) s, \( T = 0.1 \) s, and \( T = 2 \) s. The results demonstrate that for \( T = 0.03 \) s, the predicted trajectory deviates significantly from the ground truth, aligning with the poor performance reflected in Table~\ref{tab:rmse_results}. This is also consistent with the findings in Section~\ref{sec:results}, where even worse performance was observed when evaluating on noisy measurements from the perception pipeline.

\section{Complementary Evaluation Plots}
\begin{figure}[ht]
     \centering
     \includegraphics[width=0.3\textwidth]{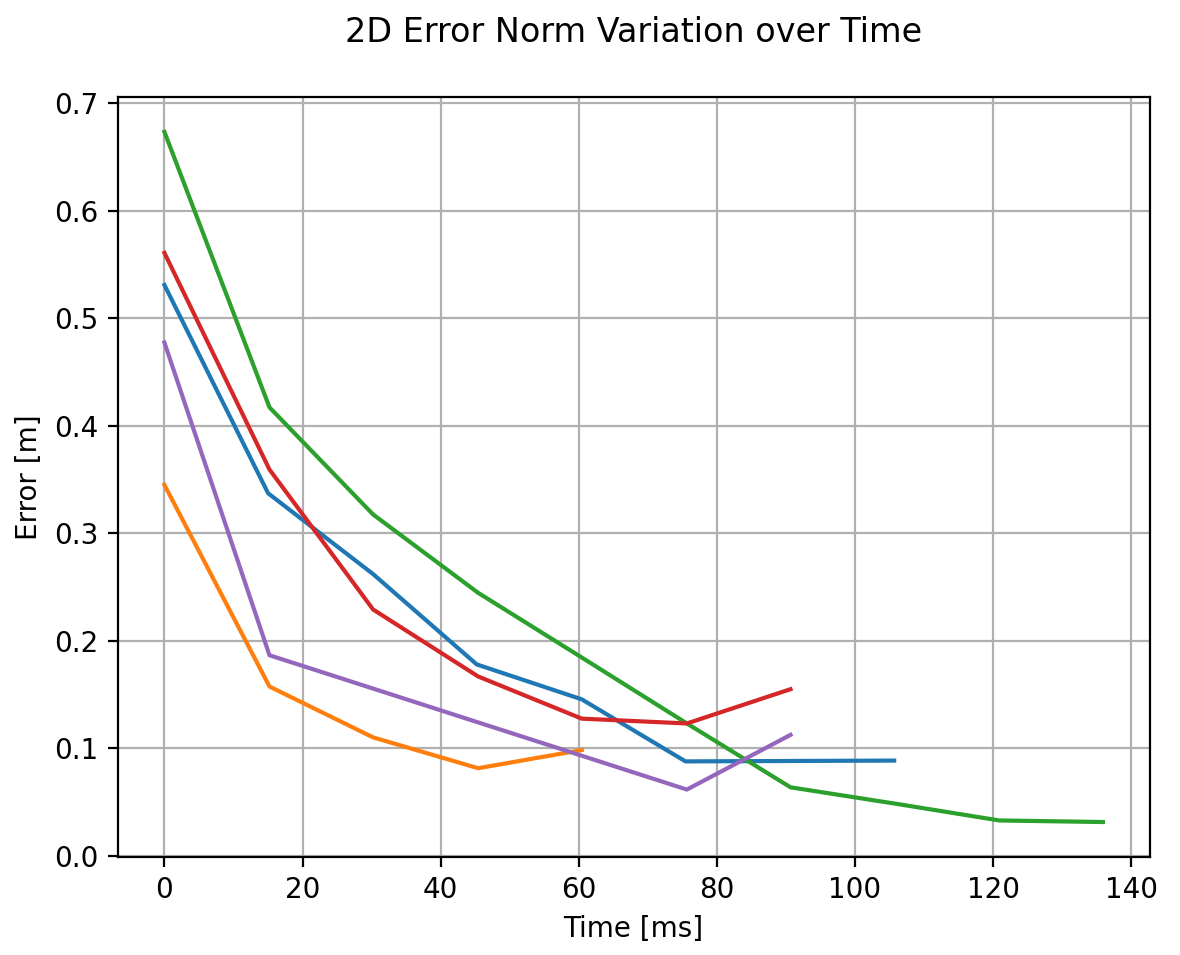}
     \caption{Error norm of the estimated bounce point on the XY plane over time for different ball trajectories. The varying lengths of the curves indicate differences in the duration for which the ball is tracked across trajectories.
     }
     \label{img:error-over-time}
\end{figure}

In this section, we provide additional plots and evaluations of the entire pipeline. First, we present error plots obtained from the online trajectory prediction method, presented in Sec.~\ref{sec:results}. Figure~\ref{img:error-over-time} highlights the gradual improvement of the trajectory prediction as the ball is continuously tracked and its path recalculated over time, using different sample trajectories. Each curve represents a different game sequence, with variations in duration due to differences in ball detectability across sequences. The results demonstrate that increasing the accumulation time window and recomputing the trajectory with more recent measurements results in a more accurate trajectory estimate.

\begin{figure}[t]
     \centering
     \includegraphics[width=0.3\textwidth]{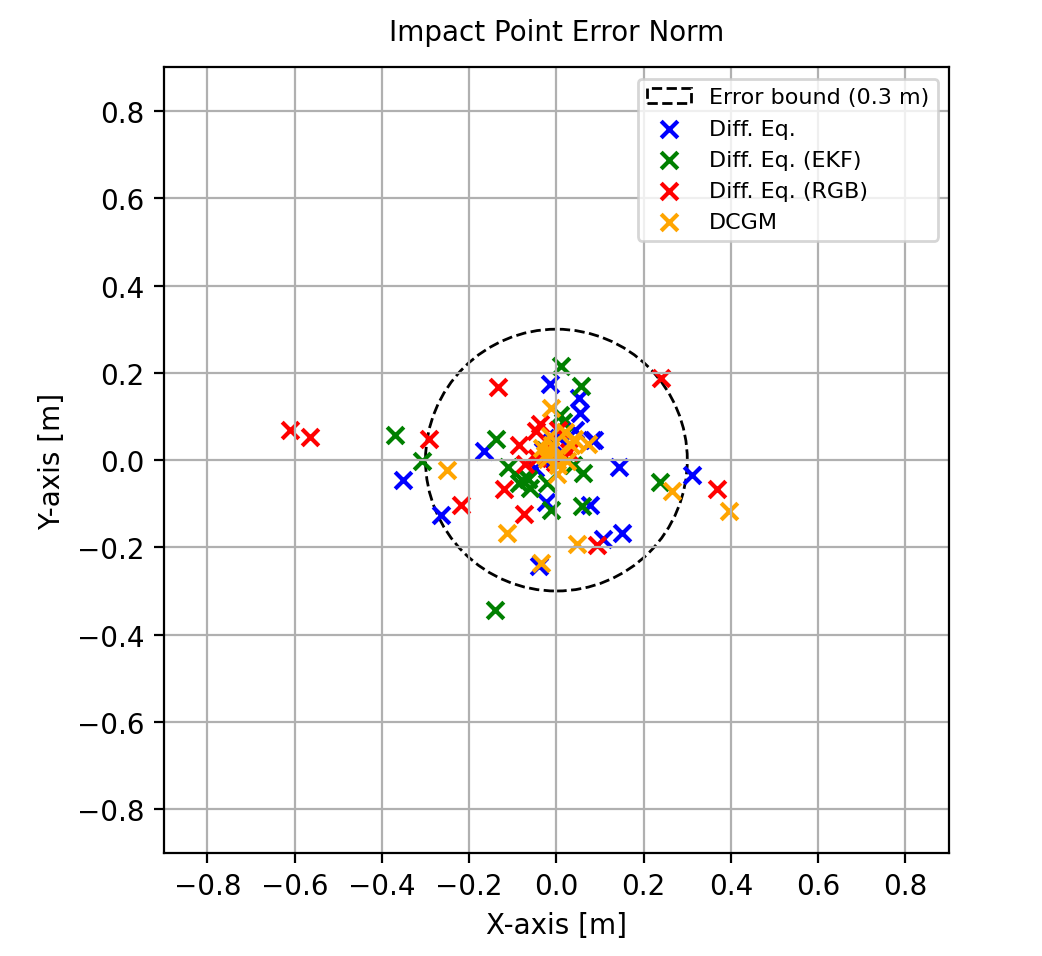}
     \caption{The plot shows the relative error of the impact point for each predicted trajectory with respect to their ground truth counterparts (each $\times$ represents an evaluated trajectory). A boundary circle with $r = 0.3$ m is shown as a reference.
     }
     \label{img:impact_points_error_norm}
\end{figure}

\vspace{2mm}
Figure~\ref{img:impact_points_error_norm}, on the other hand, shows the error distribution of the predicted bouncing points on the table compared to the ground truth counterparts. The visualization is consistent with the results in Table~\ref{tab:online-rmse-eval}, showing that the DCGM model and the standard high-frequency updates using differential equation fitting exhibit a more concentrated distribution around the origin. In contrast, the low-framerate model fitting produces a wider spread of points.

\section{Audio Signals Peak Detection}
\begin{figure*}[ht]
     \vspace{2mm}
     \centering
     \includegraphics[width=0.9\textwidth]{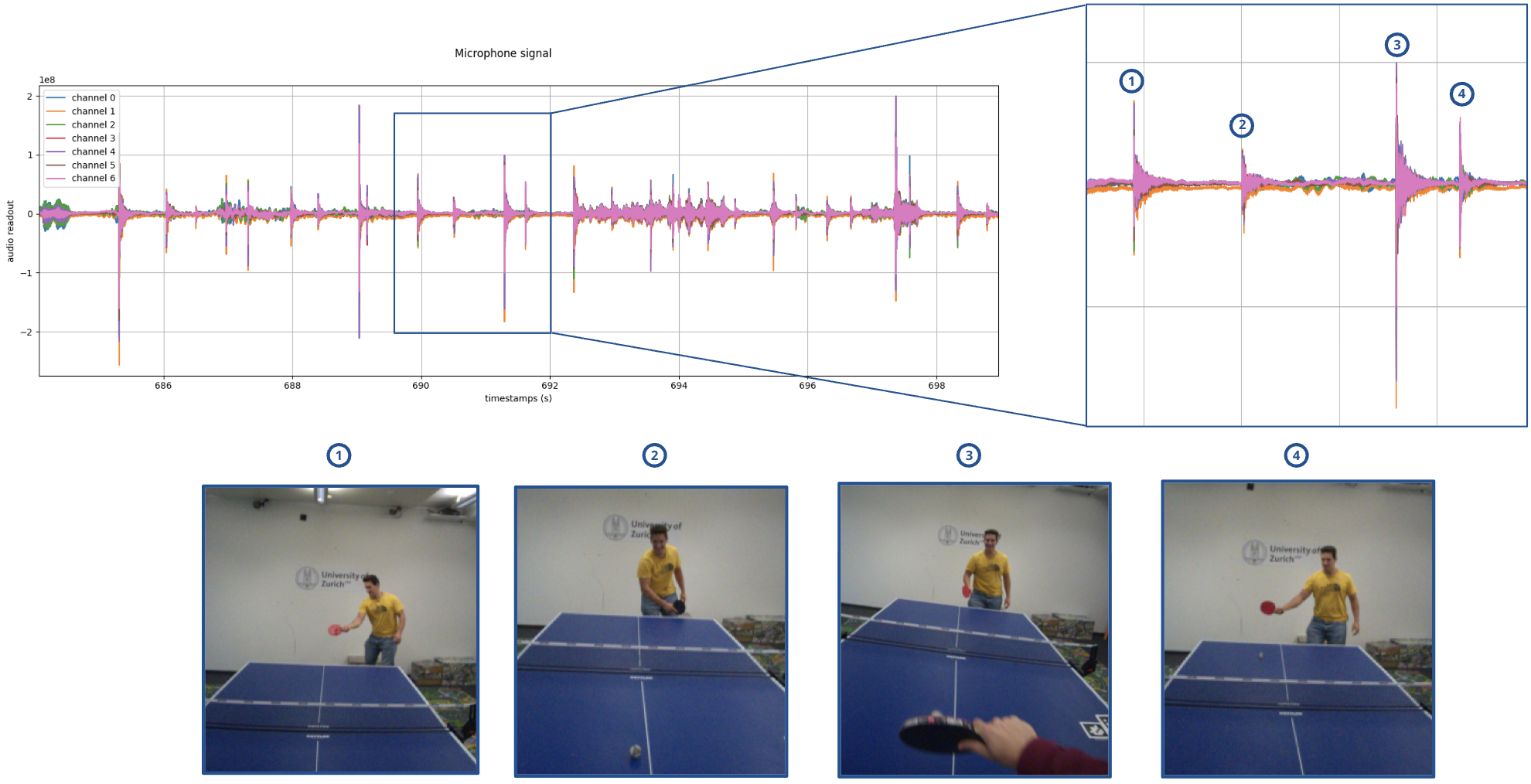}
     \caption{Microphone audio signals from a sample Project Aria glasses recording during ping-pong game.}
     \label{img:audio_peaks_game}
\end{figure*}

The evaluation of our pipeline has been carried out on sequences beginning precisely at the moment the ball impacts the opponent’s racket. To segment long game sequences, we leveraged the microphone audio signals provided by the Project Aria glasses recordings, as shown in Figure~\ref{img:audio_peaks_game}. By monitoring these signals, a pattern of peak intensities was observed, corresponding to four events: the Project Aria user hitting the ball, the ball bouncing on the user's half of the table, the ball bouncing on the opponent’s half, and the opponent hitting the ball. Each of these peaks exhibits different intensities due to their varying distances from the microphone. Specifically, the peak corresponding to the opponent’s hit has the lowest intensity. To refine our analysis, we manually filtered the audio signal using signal processing techniques.
First of all, to enhance the relevant audio features, we applied a high-pass Butterworth filter to remove low-frequency noise. After filtering, peaks in the audio signal were detected using the \texttt{find\_peaks} function of the \texttt{scipy.signal} library, after thresholding the peaks with intensity higher than $\frac{1}{4}$ of the magnitude of the time signal $y(t)$. The peak detection algorithm identifies local maxima that satisfy these conditions, ensuring that only the peaks of the opponent hitting the ball are captured. This process was repeated across multiple microphones for robust detection. For further improvements, a neural network could be trained to classify audio signals automatically. This would enable real-time detection of the opponent’s racket ball hit, optimizing computational efficiency by selectively triggering the detection pipeline.

\vspace{5mm}
\section{Comparison of Circle Fitting methods}

\begin{figure}[b]
    \centering
    \subfloat[ ]{\includegraphics[width=0.18\textwidth]{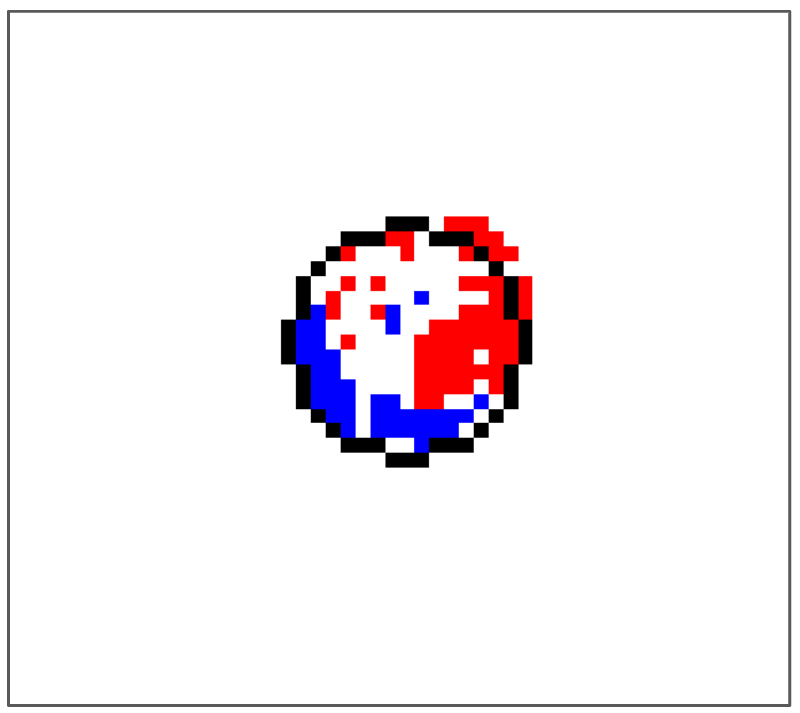}} \\
    \subfloat[ ]{\includegraphics[width=0.18\textwidth]{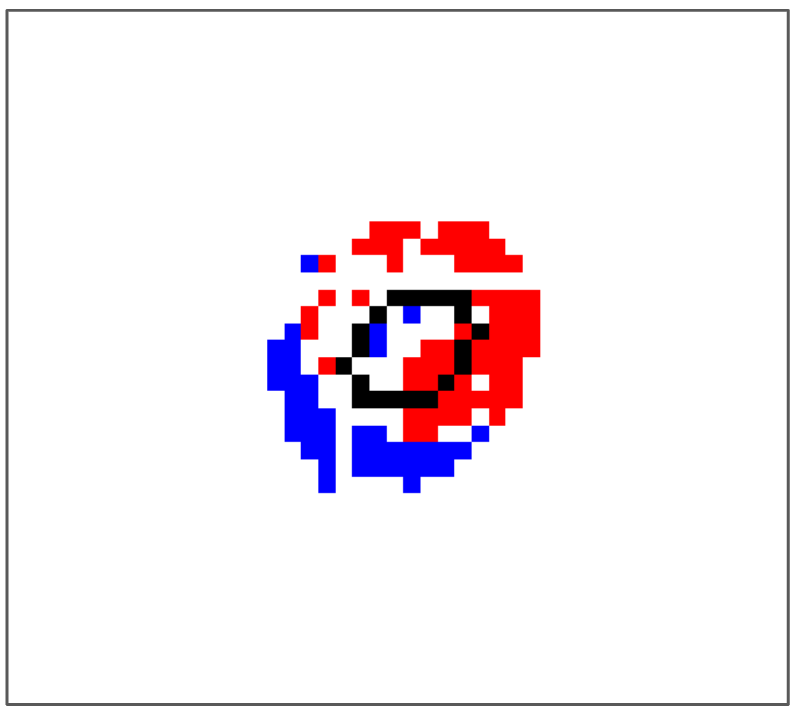}} \\
    \subfloat[ ]{\includegraphics[width=0.18\textwidth]{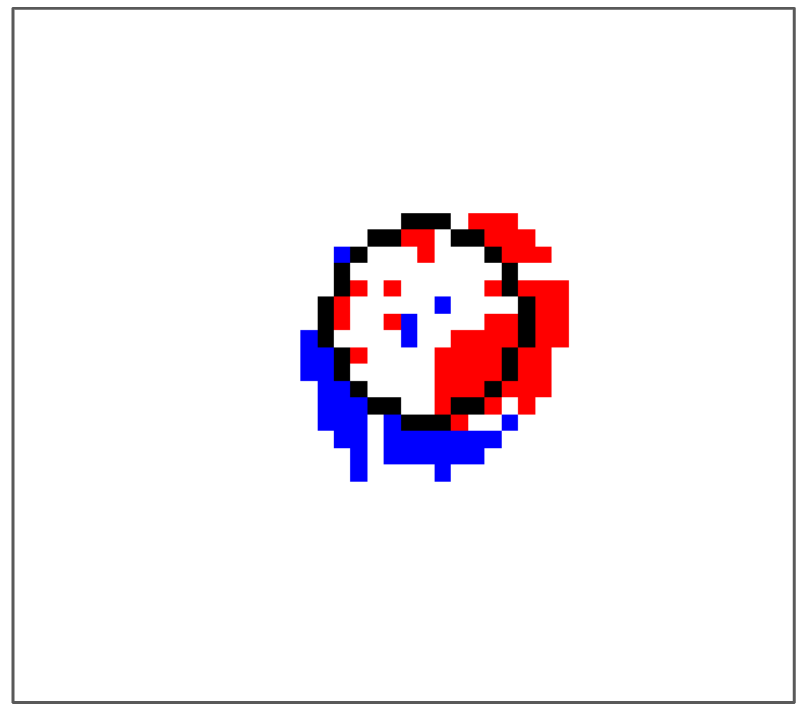}}
    \caption{Side-by-side comparison of different circle fitting methods for ball detection. (a) Our proposed method, (b) ellipse fitting, and (c) Taubin’s method. The blue and red points represent positive and negative events, respectively, while the black lines indicate the estimated radius.}
    \label{fig:circle_fitting_methods}
\end{figure}

Circle fitting is a crucial component of our algorithm, as it plays a fundamental role in estimating the depth of the ball in our monocular setup. When detecting a ping pong ball at distances of up to $3$ meters using a $640 \times 480$ resolution camera, even a one-pixel error in the estimated radius can result in a depth miscalculation of several centimeters. Since depth estimation directly influences the accuracy of the x and y coordinates, such errors can significantly impact the overall pipeline. Our proposed method, described in Section~\ref{sec:method}, is compared here with two alternative approaches: ellipse fitting and circle fitting using Taubin’s method. The ellipse fitting technique determines the mean center of the detected shape and applies Principal Component Analysis (PCA) to estimate the orientation and axis lengths. The semi-major and semi-minor axes are derived from the square root of the eigenvalues of the covariance matrix, while the orientation is dictated by the principal components. On the other hand, Taubin’s method is a geometric circle fitting approach that minimizes algebraic distance while maintaining invariance to scale transformations. Figure~\ref{fig:circle_fitting_methods} illustrates the visual results of these methods. 
Both alternative techniques tend to underestimate the ball’s radius, leading to inaccuracies in depth estimation. In contrast, our method remains the only reliable approach, ensuring consistent and precise measurements.  

\newpage

\begin{figure*}[t]
     \centering
     \includegraphics[width=0.9\textwidth]{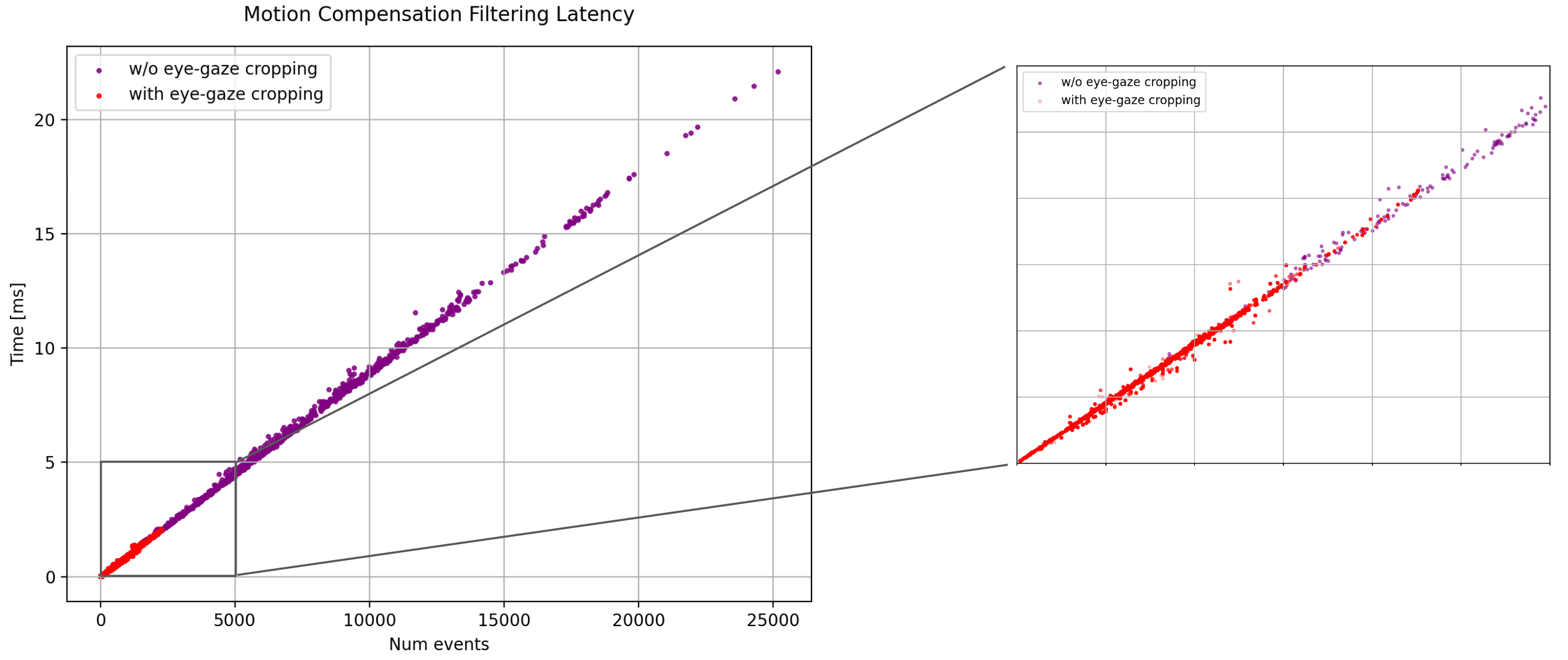}
     \caption{Time required for ego-motion compensation as a function of the number of generated events. Each dot representing a $5$ ms time window of events.}
     \label{img:motion_compensation_latency_plot}
\end{figure*}

\begin{figure*}[t]
     \centering
     \includegraphics[width=0.9\textwidth]{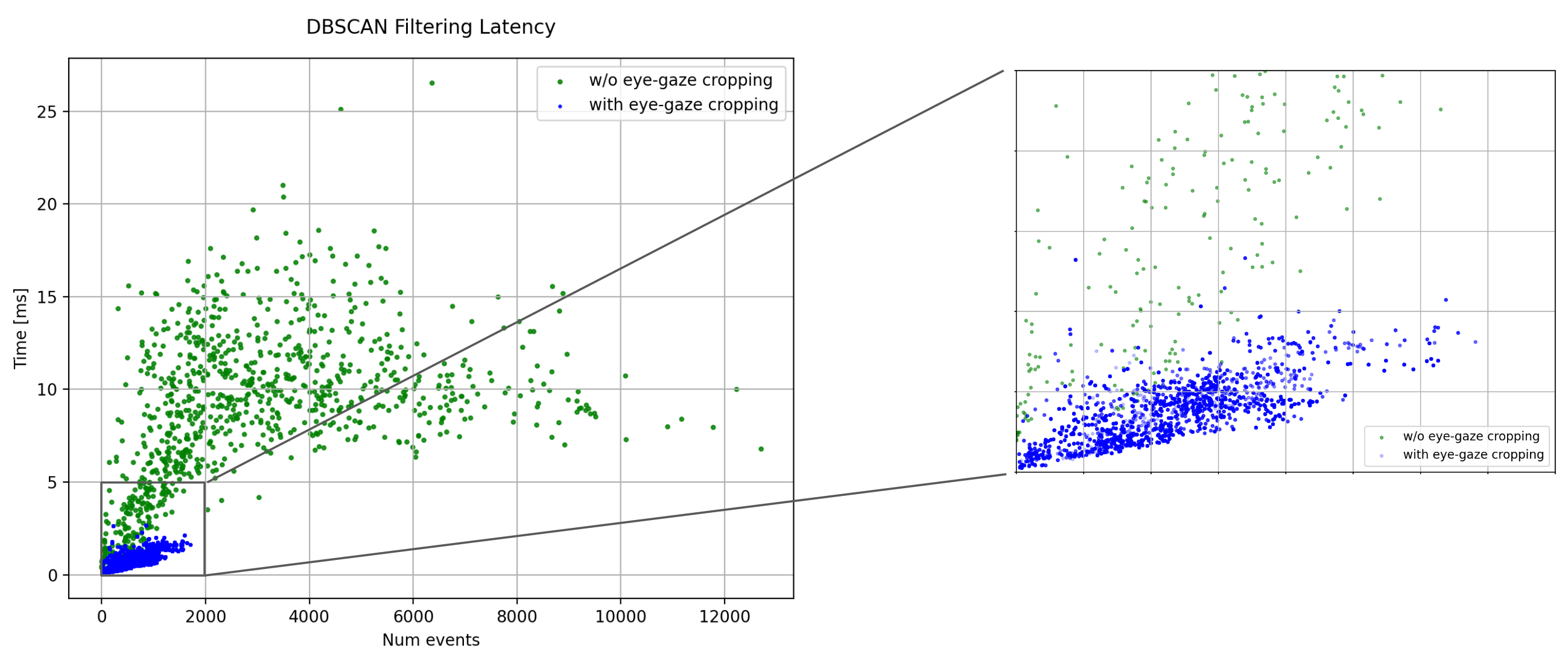}
     \caption{Time required for DBSCAN clustering of the scene's dynamic part and circularity check, based on the number of pixels from moving objects. Each dot representing a $5$ ms time window of events.}
     \label{img:dbscan_latency_plot}
\end{figure*}